\documentclass[10pt,twocolumn,letterpaper]{article}

\usepackage{wacv}
\usepackage{times}
\usepackage{epsfig}
\usepackage{graphicx}
\usepackage{amsmath}
\usepackage{amssymb}
\usepackage{booktabs}
\usepackage[accsupp]{axessibility}  

\def\wacvPaperID{119} 

\wacvapplicationstrack 

\wacvfinalcopy 

\ifwacvfinal
\usepackage[breaklinks=true,bookmarks=false]{hyperref}
\else
\usepackage[pagebackref=true,breaklinks=true,colorlinks,bookmarks=false]{hyperref}
\fi

\begin{document}

\title{Augmentation by Counterfactual Explanation - \\
Fixing an Overconfident Classifier}

\author{Sumedha Singla$^*$\\
University of Pittsburgh\\
{\tt\small sumedha.singla@pitt.edu}
\and
Nihal Murali$^*$\\
University of Pittsburgh\\
{\tt\small nihal.murali@pitt.edu}
\and
Forough Arabshahi\\
Meta AI\\
{\tt\small forough@meta.com}
\and
Sofia Triantafyllou\\
University of Crete\\
{\tt\small sof.triantafillou@gmail.com}
\and
Kayhan Batmanghelich \\
University of Pittsburgh \\
{\tt\small kayhan@pitt.edu}
}

\maketitle
\thispagestyle{empty}
\def\thefootnote{*}\footnotetext{Equal contribution}
\begin{abstract}
A highly accurate but overconfident model is ill-suited for deployment in critical applications such as healthcare and autonomous driving. The classification outcome should reflect a high uncertainty on ambiguous in-distribution samples that lie close to the decision boundary. The model should also refrain from making overconfident decisions on samples that lie far outside its training distribution, far-out-of-distribution (far-OOD), or on  unseen samples from novel classes that lie near its training distribution (near-OOD). This paper proposes an application of counterfactual explanations in fixing an over-confident classifier. Specifically, we propose to fine-tune a given pre-trained classifier using augmentations from a counterfactual explainer (ACE) to fix its uncertainty characteristics while retaining its predictive performance. We perform extensive experiments with detecting far-OOD, near-OOD, and ambiguous samples. Our empirical results show that the revised model have improved uncertainty measures, and its performance is competitive to the state-of-the-art methods.
\end{abstract}

\section{Introduction}

\newcommand{\blue}[1]{\textcolor[RGB]{0, 0, 255}{#1}}
\newcommand{\nood}[1]{\textcolor[RGB]{196, 78, 82}{#1}}

\newcommand{\aid}[1]{\textcolor[RGB]{34,135, 60}{#1}}

\newcommand{\uid}[1]{\textcolor[RGB]{76,114, 176}{#1}}

\newcommand{\food}[1]{\textcolor[RGB]{221, 132, 82}{#1}}
Deep neural networks (DNN) are increasingly being used in \emph{decision-making} pipelines for real-world high-stake applications such as medical diagnostics~\cite{esteva2017dermatologist} and autonomous driving~\cite{Autonomous}. For optimal decision making, the DNN should produce accurate predictions as well as quantify uncertainty over its predictions~\cite{Gal2016UncertaintyID,christian2017}. While substantial efforts are made to engineer highly accurate architectures~\cite{huang2017densely}, many existing state-of-the-art DNNs do not capture the uncertainty correctly~\cite{pmlr-v48-gal16}.

\begin{figure*}[htbp]
  \centering
  \includegraphics[width=0.97\linewidth]{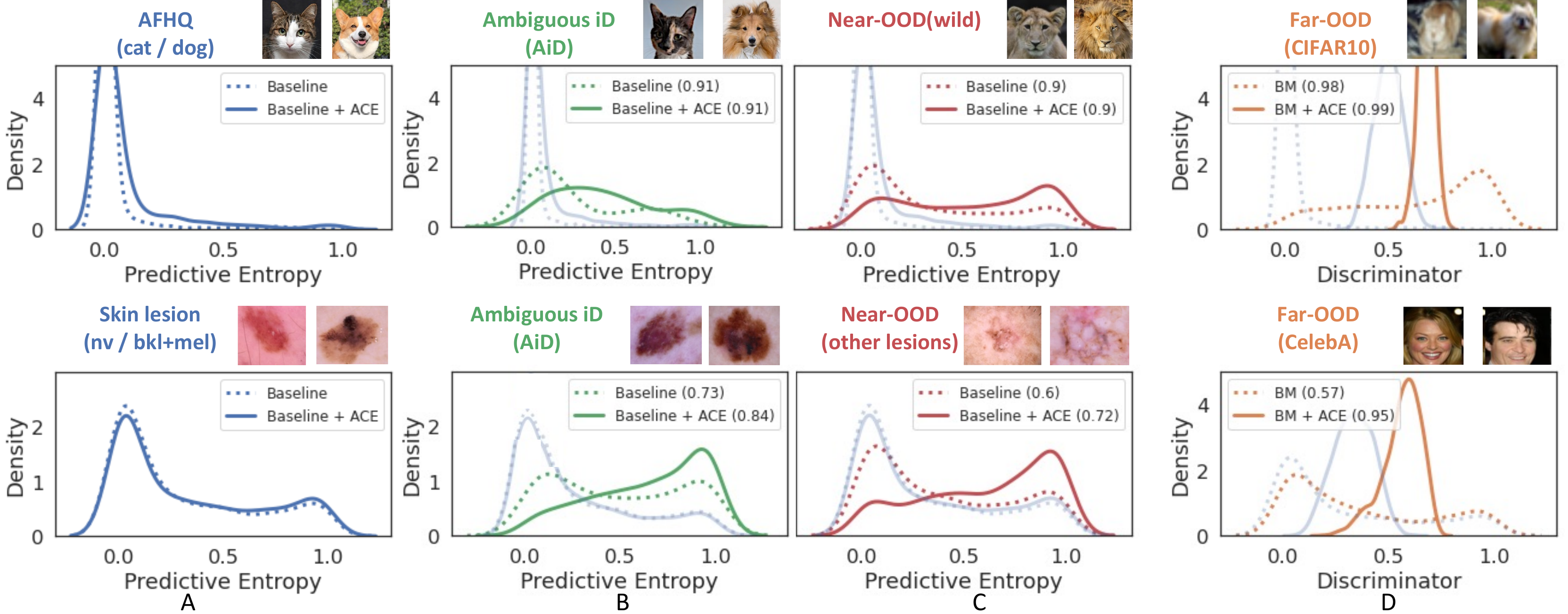}
  \caption{   \footnotesize Comparison of the uncertainty estimates from the baseline, before (dotted line) and after (solid line) fine-tuning with augmentation by counterfactual explanation (ACE). The plots visualize the distribution of predicted entropy (columns A-C) from the classifier and density score from the discriminator (column D). The y-axis of this density plot is the probability density function whose value is meaningful only for relative comparisons between groups, summarized in the legend. \textbf{A)} visualizes the impact of fine-tuning on the \textbf{\uid{in-distribution (iD)}} samples. A large overlap suggests minimum changes to classification outcome for iD samples. Next columns visualize change in the distribution for \textbf{\aid{ambiguous iD (AiD)}} \textbf{(B)} and \textbf{\nood{near-OOD}} samples \textbf{(C)}. The peak of the distribution for AiD and near-OOD samples shifted right, thus assigning higher uncertainty and reducing overlap with iD samples. \textbf{D)} compares the density score from discriminator for \textbf{\uid{iD}} (blue solid) and \textbf{\food{far-OOD}} (orange solid) samples. The overlap between the distributions is minimum, resulting in a high AUC-ROC for binary classification over uncertain samples and iD samples.  Our method improved the uncertainty estimates across the spectrum.  }
  \label{fig:all}
\end{figure*}

We consider two types of uncertainty: \emph{epistemic uncertainty}, caused due to limited data and knowledge of the model, and \emph{aleatoric uncertainty}, caused by inherent noise or ambiguity in the data~\cite{KIUREGHIAN2009105}.  We evaluate these uncertainties with respect to three test  distributions (\emph{see} Fig~\ref{fig:all}):

\begin{itemize}
\item \aid{\bf{Ambiguous in-Distribution (AiD)}}: These are the samples within the training distribution that have an inherent ambiguity in their class labels. Such ambiguity represents high aleatoric uncertainty arising from class overlap or noise~\cite{smith2018understanding}, \eg an image of a `5' that is similar to a `6'.

\item \nood{\bf{Near-OOD}}: Near-OOD represents a label shift where label
space is different between ID and OOD data.  It has high epistemic uncertainty arising from the classifier's limited information on unseen data. We use samples from unseen classes of the training distribution as near-OOD.

\item \food{\bf{Far-OOD}}: Far-OOD represents data distribution that is significantly different from the training distribution. It has high epistemic uncertainty arising from mismatch between different data distributions.

\end{itemize}

Much of the earlier work focuses on threshold-based detectors that use information from a pre-trained DNN to identify OOD samples~\cite{temperature_scaling,hendrycks17baseline,huang2021on,wang2021can,9156473}. Such methods focus on far-OOD detection and often do not address the over-confidence problem in DNN. In another line of
research, variants of Bayesian models~\cite{neal2012bayesian,pmlr-v48-gal16} and ensemble learning~\cite{Snapshot,NIPS2017_9ef2ed4b} were explored to provide reliable uncertainty estimates. Recently, there is a shift towards designing generalizable DNN that provide robust uncertainty estimates in a single forward pass~\cite{van2021feature,chen2021atom,mukhoti2021deterministic}. Such methods usually propose changes to the DNN architecture~\cite{sun2021react}, training procedure~\cite{zhang2017mixup} or loss functions~\cite{Mukhoti2020CalibratingDN} to encourage separation between ID and OOD data. Popular methods include, training deterministic DNN with a distance-aware feature space~\cite{van2020uncertainty,Liu2020SimpleAP} and 
regularizing DNN training with a generative model that simulates OOD data~\cite{lee2018training}. However, these methods require a DNN model to be trained from scratch and are not compatible with an existing pre-trained DNN. Also, they may use auxiliary data to learn to distinguish OOD inputs~\cite{liu2020energy}.

Most of the DNN-based classification models are trained to improve accuracy on a test set. Accuracy only captures the proportion of samples that are on the correct side of the decision boundary. However, it ignores the relative distance of a sample from the decision boundary~\cite{krishnan2020improving}. Ideally, samples closer to the boundary should have high uncertainty. The actual predicted value from the classifier should reflect this uncertainty via a low confidence score~\cite{Hllermeier2021AleatoricAE}. Conventionally, DNNs are trained on hard-label datasets to minimize a negative log-likelihood (NLL) loss.  Such models tend to over-saturate on NLL and end-up learning very sharp decision boundaries~\cite{guo2017calibration,mukhoti2020calibrating}. The resulting classifiers extrapolate over-confidently on ambiguous, near boundary samples, and the problem amplifies as we move to OOD regions~\cite{Gal2016UncertaintyID}.

In this paper, we propose to mitigate the overconfidence problem of a pre-trained DNN by fine-tuning it with augmentations derived from a counterfactual explainer  (ACE). We derived counterfactuals using a progressive counterfactual explainer (PCE) that create a series of perturbations of an input image, such that the classification decision is changed to a different class \cite{singla2019explanation,explaining_in_style}. PCE is trained to generate on-manifold samples in the regions between the classes. These samples along with soft labels that mimics their distance from the decision boundary, are used to fine-tuned the classifier. We hypothesis that fine-tuning on such data would broaden the classifier's decision boundary.  Our empirical results show the fine-tuned classifier exhibits better uncertainty quantification over ambiguous-iD and OOD samples.
Our contributions are as follows: (1)
We present a novel strategy to fine-tune an existing \textit{pre-trained} DNN using ACE, to improve its uncertainty estimates. (2) We proposed a refined architecture to generate counterfactual explanations that takes into account continuous condition and multiple target classes. (3) We used the discriminator of our GAN-based counterfactual explainer as a selection function to reject far-OOD samples. (4) The fine-tuned classifier with rejection head, successfully captures uncertainty over ambiguous-iD and OOD samples, and also exhibits better robustness to popular adversarial attacks.

\begin{figure*}[!h]
  \centering
  \includegraphics[width=0.8\linewidth]{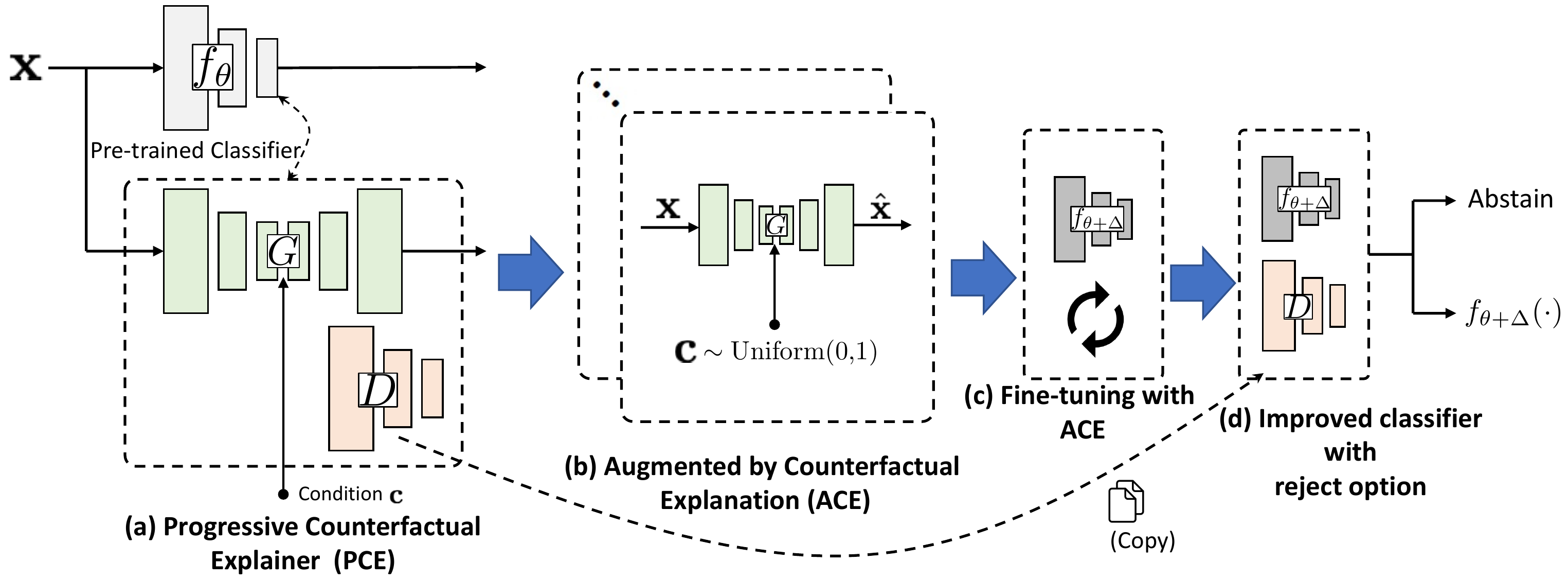}
 
  \caption{ \footnotesize (a) Given a \emph{pre-trained} classifier $f_{\theta}$, we learn a c-GAN based progressive counterfactual explainer (PCE) $G(\mathbf{x}, \mathbf{c})$, while keeping $f_{\theta}$ fixed. (b) The trained PCE creates counterfactually augmented data. (c) A combination of original training data and augmented data is used to fine-tune the classifier, $f_{\theta + \Delta}$. (d) The discriminator from PCE serves as a selection function to detect and reject OOD data.   }
  \label{fig:2}
\end{figure*}

\section{Method}

In this paper, we consider a pre-trained DNN classifier, $f_{\theta}$, with good prediction accuracy but sub-optimal uncertainty estimates. We assume $f_{\theta}$ is a differentiable function and we have access to its gradient with respect to the input, $\nabla_\mathbf{x}f_{\theta}(\mathbf{x})$, and to its final prediction outcome $f_{\theta}(\mathbf{x})$.  We also assume access to either the training data for $f_{\theta}$, or an equivalent dataset with competitive prediction accuracy. 
We further assume that the training dataset for $f_{\theta}$ has hard labels $\{0,1\}$ for all the classes. 

Our goal is to improve the pre-trained classifier  $f_{\theta}$ such that the revised model provides better uncertainty estimates, while retaining its original predictive accuracy. To enable this, we follow a two step approach. First, we  \emph{fine-tune} $f_{\theta}$ on counterfactually augmented data. The fine-tuning helps in widening the classification boundary of $f_{\theta}$, resulting in improved uncertainty estimates on ambiguous and near-OOD samples. Second, we use a density estimator to identify and reject far-OOD samples. 

We adapted previously proposed PCE~\cite{singla2019explanation} to generate counterfactually augmented data. We improved the existing implementations of PCE, by adopting
 a StyleGANv2-based backbone for the conditional-GAN in PCE. This allows using continuous vector $f_{\theta}(\mathbf{x})$ as condition for conditional generation. Further, we used the discriminator of cGAN as a \emph{selection function} to abstain revised $f_{\theta + \Delta}$ from making prediction on far-OOD samples (\textit{see} Fig.~\ref{fig:2}).

\paragraph{Notation:} The classification function is defined as $f_{\theta}: \mathbb{R}^d \rightarrow \mathbb{R}^K$, where $\theta$ represents model parameters. The training dataset for $f_{\theta}$ is defined as $\mathcal{D} = \{\mathcal{X}, \mathcal{Y}\}$, where $\mathbf{x} \in \mathcal{X} $ represents an input space and  $y \in \mathcal{Y} = \{1,2, \cdots, K\}$ is a label set over $K$ classes.  The classifier produces point estimates to approximate the posterior probability $\mathbb{P}(y|\mathbf{x}, \mathcal{D})$. 


\subsection{Progressive Counterfactual Explainer (PCE) }
\label{sec:AN}
 We designed the PCE network to take a query image ($\mathbf{x} \in \mathbb{R}^d$) and a desired classification outcome ($\mathbf{c} \in \mathbb{R}^K$) as input, and create a perturbation of a query image ($\hat{\mathbf{x}}$) such that $f_{\theta}(\hat{\mathbf{x}}) \approx \mathbf{c}$. Our formulation, $\hat{\mathbf{x}}= G(\mathbf{x}, \mathbf{c})$ allows us to use $\mathbf{c}$ to traverse through the decision boundary of $f_{\theta}$ from the original class to a counterfactual class. Following previous work~\cite{explaining_in_style,singla2019explanation,singla2021explaining}, we design the PCE to satisfy the following three properties:
 
 \begin{enumerate}
     \item \textbf{Data consistency: } The perturbed image, $\hat{\mathbf{x}}$ should be realistic and should resemble samples in $\mathcal{X}$.
     \item \textbf{Classifier consistency:} The perturbed image, $\hat{\mathbf{x}}$ should produce
the desired output from the classifier $f_{\theta}$ \ie $f_{\theta}(G(\mathbf{x}, \mathbf{c})) \approx \mathbf{c}$.

\item  \textbf{Self consistency:} Using the original classification decision $f_{\theta}(\mathbf{x})$ as condition, the PCE should produce a perturbation that is very similar to the query image, \ie $G(G(\mathbf{x}, \mathbf{c}), f_{\theta}(\mathbf{x})) = \mathbf{x}$ and $G(\mathbf{x}, f_{\theta}(\mathbf{x})) = \mathbf{x}$.
 \end{enumerate}
 \emph{Data Consistency:} 
 We formulate the PCE as a cGAN that learns the underlying data distribution of the input space $\mathcal{X}$ without an explicit likelihood assumption. The GAN model comprised of two networks -- the generator $G(\cdot)$ and the discriminator $D(\cdot)$. The $G(\cdot)$ learns to generate fake data, while the $D(\cdot)$  is trained to distinguish between the real and fake samples.  We jointly train $G, D$ to optimize the following logistic adversarial loss~\cite{goodfellow2014generative},
\begin{equation}
\begin{split}
    \mathcal{L}_{\text{adv}}(D,G) = \mathbb{E}_{\mathbf{x}}[\log D(\mathbf{x}) + \log (1 - D(G(\mathbf{x}, \mathbf{c})))]
\end{split}
\end{equation}

The earlier implementations of PCE~\cite{singla2019explanation}, have a hard constraint of representing the condition $\mathbf{c}$ as discrete variables. $f_{\theta}(\mathbf{x})$ is a continuous variable in range $[0,1]$. We adapted StyleGANv2~\cite{abdal2019image2stylegan} as the backbone of the cGAN. This formulation allow us to use $\mathbf{c} \in \mathbb{R}^K$ as condition.

 \begin{figure}[h]
  \centering
  \includegraphics[width=0.8\linewidth]{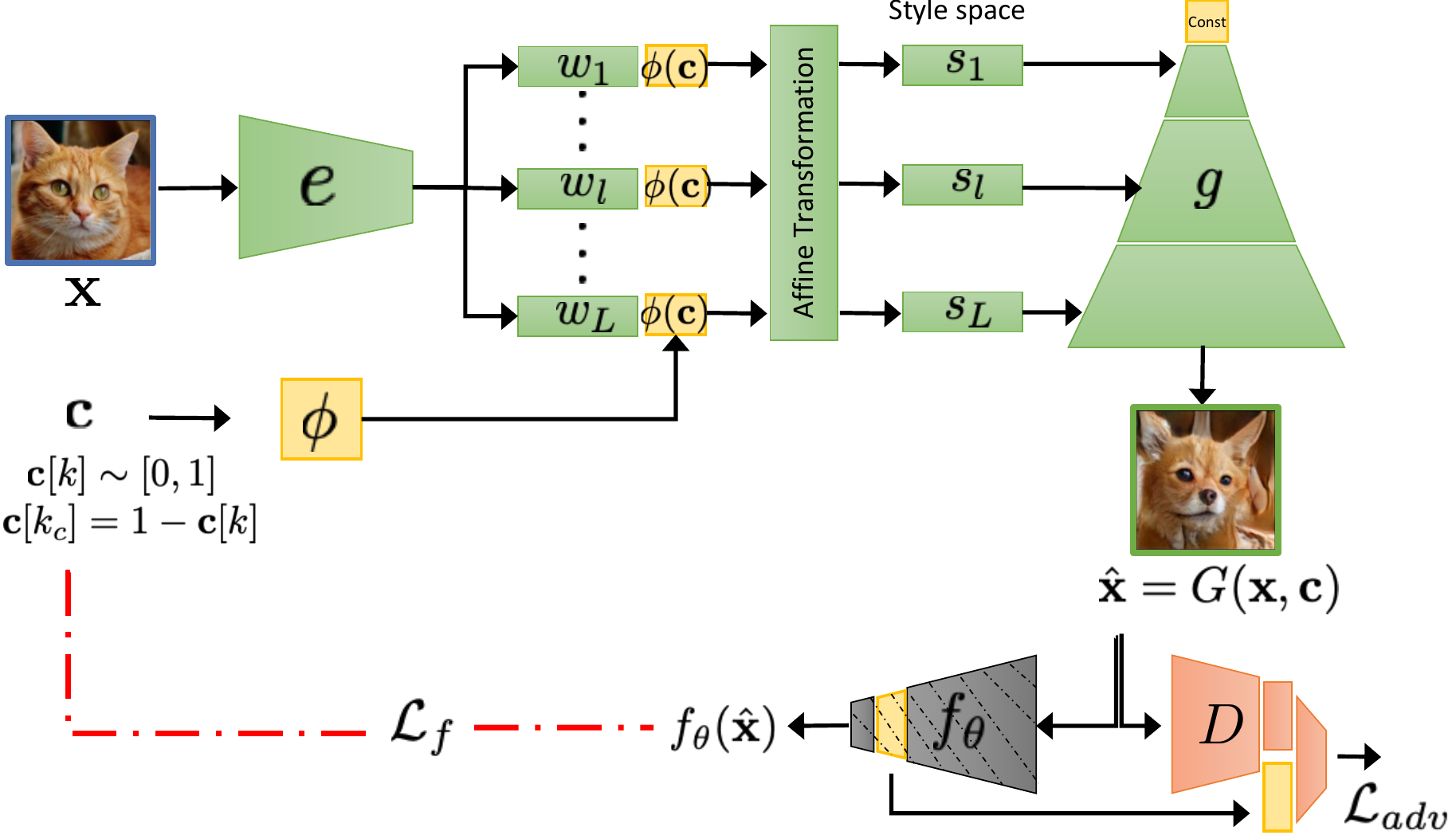}
  \caption{\footnotesize PCE: The encoder-decoder architecture to create counterfactual augmentation for a given query image. ACE: Given a query image, the trained PCE generates a series of perturbations that gradually traverse the decision boundary of $f_{\theta}$ from the original class to a counter-factual class,  while still remaining plausible and realistic-looking.    }
  \label{fig:PCE}
\end{figure}

We formulate the generator as $G(\mathbf{x}, \mathbf{c}) = g(e(\mathbf{x}), \mathbf{c})$, a composite of two functions, an image encoder $e(\cdot)$ and a conditional decoder $g(\cdot)$~\cite{abdal2019image2stylegan}. The encoder function $e: \mathcal{X} \rightarrow \mathcal{W}^+$, learns a mapping from the input space $\mathcal{X}$ to an extended latent space $\mathcal{W}^+$. The detailed architecture is provided in Fig.~\ref{fig:PCE}. 
Further, we also extended the discriminator network $D(\cdot)$ to have auxiliary information from the classifier $f_{\theta}$. Specifically, we concatenate the penultimate activations from the $f_{\theta}(\mathbf{x})$ with the penultimate activations from the $D(\mathbf{x})$, to obtain a revised representation  before the final fully-connected layer of the discriminator.
The detailed architecture is summarized in supplementary material (SM).

 We also borrow the concept of path-length regularization $\mathcal{L}_{\text{reg}}(G)$ from StyleGANv2 to enforce smoother latent space interpolations for the generator. $\mathcal{L}_{\text{reg}}(G) = \mathbb{E}_{\mathbf{w}\sim e(\mathbf{x}), \mathbf{x}\sim \mathcal{X}} (||J^T_\mathbf{w}\mathbf{x}||_2 - a)^2$, where $\mathbf{x}$ denotes random images from the training data, $J_{\mathbf{w}}$ is the Jacobian matrix, and $a$ is a constant that is set dynamically during optimization. 
\newline
\newline
\emph{Classifier consistency: }By default, GAN training is independent of the classifier $f_{\theta}$. We add a classifier-consistency loss to regularize the generator and ensure that the actual classification outcome for the generated image $\hat{\mathbf{x}}$, is similar to the condition $\mathbf{c}$ used for generation. We enforce classification-consistency by a
KullbackLeibler (KL) divergence loss as follow\cite{singla2019explanation,singla2021explaining},
\begin{equation}
    \mathcal{L}_{f}(G) = D_{KL}(f_{\theta}(\hat{\mathbf{x}})|| \mathbf{c})
\end{equation}
\emph{Self consistency: }We define the following reconstruction loss to regularize and constraint the Generator to preserve maximum information between the original image $\mathbf{x}$ and its reconstruction $\bar{\mathbf{x}}$,

\begin{equation}
    \mathcal{L}(\mathbf{x}, \bar{\mathbf{x}}) = ||\mathbf{x} - \bar{\mathbf{x}}||_1 + ||e(\mathbf{x}) - e(\bar{\mathbf{x}})||_1 
\end{equation}

Here, first term is an L1 distance loss between the input and the reconstructed image, and the second term is a style reconstruction L1 loss adapted from StyleGANv2~\cite{abdal2019image2stylegan}. We minimize this loss to satisfy the identify constraint on self reconstruction using $\bar{\mathbf{x}}_{self} = G(\mathbf{x}, f_{\theta}(\mathbf{x}))$. We further insure that the PCE learns a reversible perturbation by recovering the original image from a given perturbed image $\hat{\mathbf{x}}$ as  $\bar{\mathbf{x}}_{\text{cyclic}} = G(\hat{\mathbf{x}}, f_{\theta}(\mathbf{x}))$, where $\hat{\mathbf{x}} = G(\mathbf{x}, \mathbf{c})$ with some condition $\mathbf{c}$. Our final reconstruction loss is defined as,
\begin{equation}
    \mathcal{L}_{\text{rec}}(G) = \mathcal{L}(\mathbf{x}, \bar{\mathbf{x}}_{\text{self}}) + \mathcal{L}(\mathbf{x}, \bar{\mathbf{x}}_{\text{cyclic}})
\end{equation}
\emph{Objective function: }Finally, we trained our model in an end-to-end fashion to learn parameters for the two networks,  while keeping the classifier $f_{\theta}$ fixed. Our overall objective function is

\begin{equation}
\begin{split}
    \min_{G} \max_{D} \lambda_{\text{adv}}\left( \mathcal{L}_{\text{adv}}(D,G) +     \mathcal{L}_{\text{reg}}(G)\right) \\
    + \lambda_{f} \mathcal{L}_f(G) + \lambda_{\text{rec}} \mathcal{L}_{\text{rec}}(G),
\end{split}
\label{eq:5}
\end{equation}

where, $\lambda$'s are the hyper-parameters to balance each of the loss terms.

\subsection{Augmentation by Counterfactual Explanation}
\label{sec:ACE}
Given a query image $\mathbf{x}$, the trained PCE generates a series of  perturbations of $\mathbf{x}$ that gradually traverse the decision boundary of $f_{\theta}$ from the original class to a counterfactual class, while still remaining plausible and realistic-looking. 
We modify $\mathbf{c}$ to represent different steps in this traversal.  We start from a high data-likelihood region for original class $k$ ($\mathbf{c}[k] \in [0.8, 1.0]$), walk towards the decision hyper-plane ($\mathbf{c}[k] \in [0.5, 0.8)$), and eventually cross the decision boundary ($\mathbf{c}[k] \in [0.2, 0.5)$) to end the traversal in a high data-likelihood region for the counterfactual class $k_c$ ($\mathbf{c}[k] \in [0.0, 0.2)$). Accordingly, we  set $\mathbf{c}[k_{c}] = 1- \mathbf{c}[k]$.  

Ideally, the predicted confidence from NN should be  indicative of the distance from the decision boundary. Samples that lies close to the decision boundary should have low confidence, and confidence should increase as we move away from the decision boundary. We used $\mathbf{c}$ as a pseudo indicator of confidence to generate synthetic augmentation. Our augmentations are essentially showing how the query image $\mathbf{x}$ should be modified to have low/high confidence.

To generate counterfactual augmentations, we randomly sample a subset of real training data as $\mathcal{X}_r {\displaystyle \subset }\mathcal{X}$. Next, for each $\mathbf{x} \in \mathcal{X}_r$ we generate multiple augmentations ($\hat{\mathbf{x}} = G(\mathbf{x}, \mathbf{c})$) by randomly sampling $\mathbf{c}[k] \in [0,1]$. We used $\mathbf{c}$ as soft label for the generate sample while fine-tuning the $f_{\theta}$. The  $\mathcal{X}_{c}$ represents our pool of generated augmentation images. Finally, we create a new dataset by randomly sampling images from $\mathcal{X}$ and $\mathcal{X}_c$. We fine-tune the $f_{\theta}$ on this new dataset, for only a few epochs, to obtain a revised classifier  given as $f_{\theta + \Delta}$. In our experiments, we show that the revised decision function $f_{\hat{\theta}}$ provides improved confidence estimates for AiD and near OOD samples and demonstrate robustness to adversarial attacks, as compared to given classifier $f_{\theta}$.

\subsection{Discriminator as a Selection Function}
\label{sec:SC}
A selection function $g: \mathcal{X} \rightarrow \{0,1\}$ is an addition head on top of a classifier that decides when the classifier should  abstain from making a prediction. We propose to use the discriminator network $D(\mathbf{x})$ as a selection function for $f_{\theta}$. Upon the convergence of the PCE training, the generated samples resemble the in-distribution training data. Far-OOD samples are previously unseen samples which are very different from the training input space. Hence, $D(\cdot)$ can help in detecting such samples.  Our final improved classification function is represented as follow,

\begin{equation}
    (f,D)(\mathbf{x})= 
\begin{cases}
    f_{\theta + \Delta} (\mathbf{x}),& \text{if } D(\mathbf{x}) \geq h\\
    \texttt{Abstain},              & \text{otherwise}
\end{cases}
\end{equation}

where, $f_{\theta + \Delta}$ is the fine-tuned classifier and $D(\cdot)$ is a discriminator network from the PCE which serves as a selection function that permits $f$ to make prediction if $D(\mathbf{x})$ exceeds a threshold $h$ and abstain otherwise.


\section{Related Work}

\textbf{Uncertainty estimation in pre-trained DNN models:} Much of the prior work focused on deriving uncertainty measurements from a pre-trained DNN output~\cite{hendrycks17baseline,temperature_scaling,liang2018enhancing,liu2020energy}, feature representations~\cite{lin2021mood,Lee2018ASU} or gradients~\cite{huang2021on}. Such methods use a threshold-based scoring function to identify OOD samples. The scoring function is derived from softmax confidence scores~\cite{hendrycks17baseline}, scaled logit~\cite{temperature_scaling,lin2021mood}, energy-based scores~\cite{liu2020energy,wang2021can} or gradient-based scores~\cite{huang2021on}. These methods help in identifying OOD samples but did not address the over-confidence problem of DNN, that made identifying OOD non-trivial in the first place~\cite{Hein2019WhyRN,nguyen2015deep}. We propose to mitigate the over-confidence issue by fine-tuning the pre-trained classifier using ACE. Further, we used a hard threshold on the density score provided by the discriminator of the GAN-generator,  to identify OOD samples. 

\textbf{DNN designs for improved uncertainty estimation:} 
The Bayesian neural networks are the gold standard for reliable uncertainty quantification~\cite{neal2012bayesian}. Multiple approximate Bayesian approaches have been proposed to achieve tractable inference and to reduce computational complexity~\cite{NIPS2011_7eb3c8be,Weight_Uncertainty,NIPS2015_bc731692,pmlr-v48-gal16}.  Popular non-Bayesian methods include deep ensembles~\cite{NIPS2017_9ef2ed4b} and their variant~\cite{Snapshot,loss_surface}. However, most of these methods are computationally expensive and requires multiple passes during inference. An alternative approach is to modify DNN training~\cite{label_smoothing_szegedy,zhang2017mixup,manifold_mixup}, loss function~\cite{Mukhoti2020CalibratingDN}, architecture~\cite{sun2021react,Liu2020SimpleAP,Geifman2019SelectiveNetAD} or end-layers~\cite{van2020uncertainty,9156473} to support improved uncertainty estimates in a single forward-pass. Further, methods such as DUQ~\cite{van2020uncertainty} and DDU~\cite{mukhoti2021deterministic} proposed modifications to enable the separation between aleatoric and epistemic uncertainty. Unlike these methods, our approach improves the uncertainty estimates of any existing pre-trained classifier, without changing its architecture or training procedure. We used the discriminative head of the fine-tuned classifier to capture aleatoric uncertainty and the density estimation from the GAN-generator to capture epistemic uncertainty.    

\textbf{Uncertainty estimation using GAN:} A popular technique to fix an over-confident classifier is to regularize the model with an auxiliary OOD data which is either realistic~\cite{hendrycks2018deep,Mohseni2020SelfSupervisedLF,PAPADOPOULOS2021138,chen2021atom,LiangEnhancing} or is generated using GAN~\cite{Ren2019LikelihoodRF,lee2018training,Mandal_2019_CVPR,Xiao2020LikelihoodRA,Serr2020Input}. 
Such regularization helps the classifier to assign lower confidence to anomalous samples, which usually lies in the low-density regions.
Defining the scope of OOD a-priori is generally hard and can potentially cause a selection bias in the learning. Alternative approaches resort to estimating in-distribution density~\cite{Subramanya2017ConfidenceEI}.
 Our work fixed the scope of GAN-generation to counterfactual generation. Rather than merging the classifier and the GAN training, we train the GAN in a post-hoc manner to explain the decision of an existing classifier. This strategy defines OOD in the context of pre-trained classifier’s decision boundary. Previously, training with CAD have shown to improved generalization performance on OOD samples~\cite{Kaushik2021ExplainingTE}. However, this work is limited to Natural Language Processing, and requires human intervention while curating CAD~\cite{Kaushik2020Learning}. In contrast, we train a GAN-based counterfactual explainer~\cite{singla2021explaining,explaining_in_style} to derive CAD.



\section{Experiment}

\newcommand\Tstrut{\rule{0pt}{2.6ex}}         
\newcommand\Bstrut{\rule[-1.0ex]{0pt}{0pt}}   

  \begin{figure*}[t]
  \centering
  \includegraphics[width=0.8\linewidth]{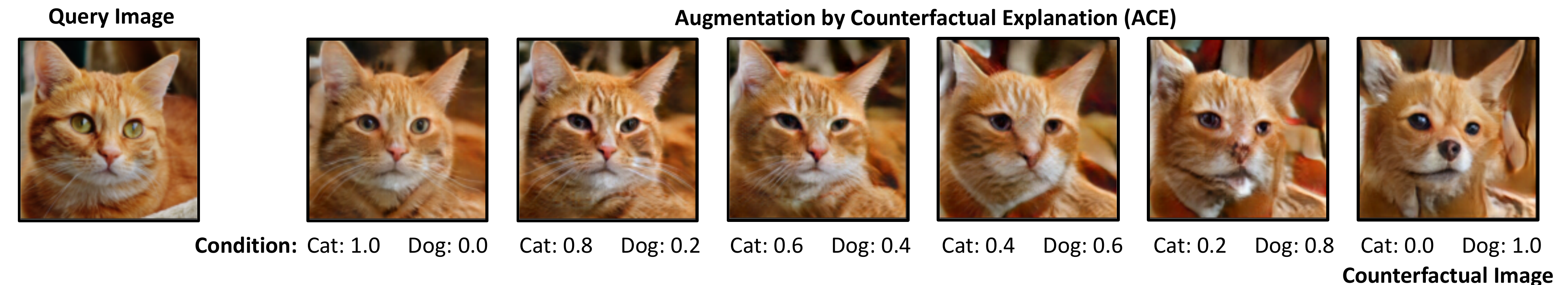}
  \caption{  \footnotesize An example of counterfactually generated data by the progressive counterfactual explainer (PCE). More examples are provided in the Supplementary Material.}
  \label{fig:skin}
\end{figure*}

 We consider four classification problems, in increasing level of difficulty: 
 \begin{enumerate}
  \item AFHQ~\cite{choi2020afhq}: We consider binary classification over well separated classes, cat vs dog.

    \item Dirty MNIST~\cite{mukhoti2021deterministic}: We consider multi-class classification over hand-written digits 0-6. The dataset is a combination of original MNIST~\cite{lecun1998mnist} and simulated samples from a variational decoder. The samples are generated by combining latent representation of different digits, to simulate ambiguous samples, with multiple plausible labels~\cite{mukhoti2021deterministic}.
    
    \item CelebA~\cite{liu2015celeba}: We consider a multi-label classification setting over `young' and `smiling' attributes. Without age labels, identifying 'young' faces is a challenging task. 
    
    \item Skin lesion (HAM10K)~\cite{tschandl2018ham10000}:  We consider a binary classification to separate Melanocytic nevus (nv) from Melanoma (mel) and Benign Keratosis (bkl) lesions.  Skin lesion classification is a challenging task as different lesions may exhibit similar features~\cite{nachbar1994abcd}. 
   
 \end{enumerate}

 \emph{Architecture details: }We consider state-of-the-art DenseNet~\cite{huang2017densely} architecture for the baseline. The \emph{pre-trained} DenseNet model followed the training procedures as described in~\cite{huang2017densely}. In order to keep the architecture and training procedure of PCE simple, we consider the default training parameters from~\cite{abdal2019image2stylegan} for training the StyleGANv2. This encourages reproducibility as we didn't do hyper-parameter tuning for each dataset and classification model. 
 For training  StyelGANv2, we use a randomly sampled subset $(\sim50\%)$ of the baseline model's training data. For multi-class classification, we consider all pairs of classes while creating counterfactual augmentations. For fine-tuning the baseline, we create a new dataset with 30\% counterfactually generated samples and 70\% real samples from the original training set. All the results are reported on the test set of the baseline. In all our experiments, we used $\lambda_{adv} = 10$, $\lambda_{rec} = 100$, $\lambda_{f} = 10$, and $h=0.5$.

 \emph{Comparison methods: } Our baseline is a standard DNN classifier $f_{\theta}$ trained with cross-entropy loss. For baseline and its post-hoc variant with temperature-scaling (\textbf{TS}), we used threshold over predictive entropy (PE) to identify OOD. PE is defined as $-\sum[f_{\theta}(\mathbf{x})]_k \log[f_{\theta}(\mathbf{x})]_k$. Next, we compared against following five methods:
 \textbf{mixup}: baseline model with mixup training using $\alpha = 0.2$~\cite{zhang2017mixup}; deterministic uncertainty quantification (\textbf{DUQ})~\cite{van2020uncertainty}: baseline model with radial basis function as end-layer; \textbf{DDU}: 
 that use the closest kernel distance to quantify uncertainties;  
 \textbf{MC Dropout} (with 20 dropout samples)~\cite{pmlr-v48-gal16}; and five independent runs of baseline as \textbf{5-Ensemble}~\cite{NIPS2017_9ef2ed4b}.  The ensemble approaches are an upper bound for UQ.

\begin{table*}[!htb]
\caption{ \footnotesize Performance of different methods on identifying \textbf{\aid{ambiguous in-distribution (AiD)}} samples. For all metrics, higher is better. The best results from the methods that require a single forward pass at inference time are highlighted.
}
\label{AiD-table}
\begin{center}
\scriptsize
    \begin{tabular}{cccccc}
    \hline
    \multicolumn{1}{c}{\bf Train }  &  \multicolumn{1}{c}{\bf Method/ }  & \multicolumn{1}{c}{\bf Test-Set} &  \multicolumn{2}{c}{\bf Identifying \aid{AiD}}  \Tstrut{}\\
    \bf Dataset & \bf Model  & Accuracy & AUC-ROC & TNR@TPR95  \Bstrut{}\\
    \hline
    
    & Baseline  &   99.44$\pm$0.02  &  0.87$\pm$0.04  & 48.93$\pm$10 \Tstrut{}\\ 
     & Baseline+TS~\cite{guo2017temperaturescaling} & 99.45$\pm$0.00  &   0.85$\pm$0.07  &  48.77$\pm$9.8 \\
     & Baseline+TS+ODIN~\cite{liang2018enhancing}  &  99.45$\pm$0.00   &  0.85$\pm$0.06  & 35.72$\pm$1.26 \\ 
     & Baseline+energy~\cite{liu2020energy}  &   99.44$\pm$0.02  &  0.87$\pm$0.06  & 49.00$\pm$1.64 \\ 
    & Mixup~\cite{zhang2017mixup} & 99.02$\pm$0.10    & 0.80$\pm$0.05  &  35.66$\pm$6.7 \\ 
    AFHQ & DUQ~\cite{van2020uncertainty} & 94.00$\pm$1.05  & 0.67$\pm$0.01    & 26.15$\pm$4.5 \\ 
    & DDU~\cite{mukhoti2021deterministic} & 97.66$\pm$1.10  &  0.74$\pm$0.02  &  19.65$\pm$4.5 \\

    & \bf Baseline+ACE & \bf 99.52$\pm$0.21  & \bf 0.91$\pm$0.02  &  \bf 50.75$\pm$3.9 \Bstrut{}\\
    \cline{2-5}
    & MC Dropout~\cite{pmlr-v48-gal16} & 98.83$\pm$1.12 & 0.87$\pm$0.04 & 51.56$\pm$1.2 \Tstrut{}\\
    & 5-Ensemble~\cite{NIPS2017_9ef2ed4b} &  99.79$\pm$0.01  &  0.98$\pm$0.01  &  51.93$\pm$2.7 \Bstrut{} \\
    \hline 
     & Baseline  &   \bf 95.68$\pm$0.02 & \bf0.96$\pm$0.00  & 28.5$\pm$2.3 \Tstrut{}\\ 
     & Baseline+TS~\cite{guo2017temperaturescaling} &\bf 95.74$\pm$0.02 &  0.94$\pm$0.01  & 27.90$\pm$1.3\\ 
     & Baseline+TS+ODIN~\cite{liang2018enhancing}  &  \bf 95.74$\pm$0.02   & 0.79$\pm$0.03   & 13.25$\pm$4.88 \\ 
     & Baseline+energy~\cite{liu2020energy}  & \bf 95.68$\pm$0.02   & 0.80$\pm$0.03   & 17.60$\pm$0.55 \\
     & Mixup~\cite{zhang2017mixup} & 94.66$\pm$0.16   &  0.94$\pm$0.02  & 25.78$\pm$2.1 \\ 
    Dirty & DUQ~\cite{van2020uncertainty} & 89.34$\pm$0.44  & 0.67$\pm$0.01   & 23.89$\pm$1.2 \\ 
    MNIST & DDU~\cite{mukhoti2021deterministic}& 93.52$\pm$1.12   &     0.65$\pm$0.12  &   20.78$\pm$4.0 \\

    & \bf Baseline+ACE &  \bf 95.36$\pm$0.45  &  0.86$\pm$0.01  &  \bf 34.12$\pm$2.6 \Bstrut{}\\
    \cline{2-5}
    & MC Dropout~\cite{pmlr-v48-gal16} & 89.50$\pm$1.90  &     0.75$\pm$0.07  &  36.10$\pm$1.8 \Tstrut{}\\ 
    & 5-Ensemble~\cite{NIPS2017_9ef2ed4b} &  95.90$\pm$0.12   &     0.98$\pm$0.02  &  34.87$\pm$3.4 \Bstrut{}\\
    
    \hline
    
    & Baseline  & \bf 89.36$\pm$0.96 & 0.73$\pm$0.01 & 17.18$\pm$1.6 \Tstrut{}\\  
    & Baseline+TS~\cite{guo2017temperaturescaling} & 89.33$\pm$0.01 & 0.72$\pm$0.02 & 17.21$\pm$1.5   \\  
    & Baseline+TS+ODIN~\cite{liang2018enhancing}  & 89.33$\pm$0.01    &0.57$\pm$0.01    & 6.34$\pm$0.38 \\
    & Baseline+energy~\cite{liu2020energy}  &  89.36$\pm$0.96   & 0.57$\pm$0.28   & 4.87$\pm$0.32 \\ 
    & Mixup~\cite{zhang2017mixup} & 89.04$\pm$0.47 & \bf 0.74$\pm$0.02 & 15.09$\pm$1.9 \\ 
    CelebA & DUQ~\cite{van2020uncertainty} & 71.75$\pm$0.01 & 0.65$\pm$0.01 & 14.20$\pm$1.0 \\ 
     & DDU~\cite{mukhoti2021deterministic} & 70.15$\pm$0.02 & 0.67$\pm$0.06 & 11.39$\pm$0.4 \\

     & \bf Baseline+ACE &  86.8$\pm$0.79 & \bf 0.74$\pm$0.06 & \bf 22.36$\pm$ 2.3 \Bstrut{}\\
    \cline{2-5}
    & MC Dropout~\cite{pmlr-v48-gal16} & 89.86$\pm$0.33 & 0.73$\pm$0.03 & 19.78$\pm$0.7 \Tstrut{}\\
    & 5-Ensemble~\cite{NIPS2017_9ef2ed4b} &  90.76$\pm$0.00 & 0.84$\pm$0.11 & 17.79$\pm$0.6 \Bstrut{}\\
    \hline
    
     & Baseline  &    85.88$\pm$0.75   & 0.82$\pm$0.06  & 20.52$\pm$3.7 \Tstrut{}\\ 
     & Baseline+TS~\cite{guo2017temperaturescaling} &  \bf 86.27$\pm$0.40  &     \bf 0.84$\pm$0.03  & 23.34$\pm$2.8 \\ 
     & Baseline+TS+ODIN~\cite{liang2018enhancing}  & 86.27$\pm$0.40    & 0.78$\pm$0.01   & 15.87$\pm$4.33 \\
     & Baseline+energy~\cite{liu2020energy}  & 85.88$\pm$0.75    & 0.77$\pm$0.12   & 18.40$\pm$0.51 \\  
    & Mixup~\cite{zhang2017mixup} &  85.81$\pm$0.61  &  \bf  0.84$\pm$0.04& 31.29$\pm$7.0 \\ 
    Skin-Lesion & DUQ~\cite{van2020uncertainty} & 75.47$\pm$5.36  &  0.81$\pm$0.02     &  30.12$\pm$4.4\\ 
    (HAM10K) & DDU~\cite{mukhoti2021deterministic}& 75.84$\pm$2.34   &     0.79$\pm$0.03  &   26.12$\pm$6.6 \\

    & \bf Baseline+ACE & 81.21$\pm$1.12  & \bf 0.84$\pm$0.05  & \bf 71.60$\pm$3.8 \Bstrut{}\\
    \cline{2-5}
    & MC Dropout~\cite{pmlr-v48-gal16} & 84.90$\pm$1.17   &  0.85$\pm$0.06  &  43.78$\pm$1.9 \Tstrut{}\\ 
    & 5-Ensemble~\cite{NIPS2017_9ef2ed4b} &  87.89$\pm$0.13   &     0.86$\pm$0.02  &  40.49$\pm$5.1 \Bstrut{}\\
    
    \hline
    \end{tabular}
\end{center}
\end{table*}

 \subsection{Identifying AiD samples}

We do not have access to ground truth labels marking the samples that are AiD. Hence, we used the PE estimates from an MC Dropout classifier to obtain pseudo-ground truth for AiD classification.  Specifically,   
 we sort the test set using PE and consider the top 5 to 10\% samples as AiD.  In Fig.~\ref{fig:all}, we qualitatively compare the PE distribution from the given baseline and its fine-tuned version (baseline + ACE). Fine-tuning resulted in minor changes to the PE distribution of the iD samples (Fig.~\ref{fig:all}.A). 
 We observe a significant separation in the PE distribution of AiD samples and the rest of the test set (Fig.~\ref{fig:all}.B), even on the baseline. This suggests that the PE correctly captures the aleatoric uncertainty.
 
 Table~\ref{AiD-table} compares our model to several baselines.  We report the test set accuracy,  the AUC-ROC for the binary task of identifying AiD samples and the  true negative rate (TNR) at 95\% true positive rate (TPR) (TNR@TPR95), which simulates
an application requirement that the recall of in-distribution
data should be 95\%~\cite{9156473}. For all metrics higher value is better. Our model outperformed other deterministic models in identifying AiD samples with a high AUC-ROC and TNR@TPR95 across all datasets.


\subsection{Detecting OOD samples}

We consider two tasks to evaluate the model’s OOD detection performance. First, a standard OOD task where OOD samples are derived from a separate dataset. Second, a difficult near-OOD detection task where OOD samples belongs to novel classes from the same dataset, which are not seen during training. We consider the following OOD datasets:

 \begin{enumerate}
 
\item AFHQ~\cite{choi2020afhq}: We consider ``wild" class from AFHQ to define near-OOD samples. For the far-OOD detection task, we use the CelebA dataset, and also cat/dog images from CIFAR10~\cite{Krizhevsky2009LearningML}.
  
    \item Dirty MNIST~\cite{mukhoti2021deterministic}: We consider digits 7-9 as near-OOD samples. For far-OOD detection, we use SVHN~\cite{37648} and fashion MNIST~\cite{Xiao2017FashionMNISTAN} datasets.

 \item CelebA~\cite{liu2015celeba}: We consider images of kids in age-group: 0-11 from the UTKFace~\cite{zhifei2017cvpr} dataset to define the near-OOD samples. For far-OOD detection task, we use the AFHQ and CIFAR10 datasets.
 
    \item Skin lesion (HAM10K)~\cite{tschandl2018ham10000}: We consider samples from lesion types: Actinic Keratoses and Intraepithelial Carcinoma (akiec), Basal Cell Carcinoma (bcc), Dermatofibroma (df) and Vascular skin lesions (vasc) as near-OOD. For far-OOD, we consider CelebA and an additional simulated dataset with different skin textures/tones.

 \end{enumerate}
 
 \begin{table*}[!htb]
\caption{ \footnotesize OOD detection performance for different baselines. \textbf{\nood{Near-OOD}} represents label shift, with samples from the unseen classes of the same dataset. \textbf{\food{Far-OOD}} samples are from a separate dataset.  The numbers are averaged over five runs. 
}
\label{OOD-table}
\begin{center}
\scriptsize
\resizebox{\textwidth}{!}{%
    \begin{tabular}{cccccccc}
    \hline 
    \multicolumn{1}{c}{\bf Train }  &  \multicolumn{1}{c}{\bf Method }  &   \multicolumn{2}{c}{\bf \nood{Near-OOD} (Wild)} &  \multicolumn{2}{c}{\bf \food{Far-OOD} (CIFAR10)} &  \multicolumn{2}{c}{\bf \food{Far-OOD} (CelebA)} \Tstrut{}\\
    \bf Dataset & \bf   & AUC-ROC & TNR@TPR95 & AUC-ROC & TNR@TPR95 &   AUC-ROC & TNR@TPR95 \Bstrut{}\\
    \hline
     & Baseline  &  0.88$\pm$0.04 & 47.40$\pm$5.2    & 0.95$\pm$0.04  & 73.59$\pm$9.4 &   0.95$\pm$0.03 & 70.69$\pm$8.9 \Tstrut{}\\
     
     & Baseline+TS~\cite{guo2017temperaturescaling} & 0.88$\pm$0.03   & 45.53$\pm$9.8 &   0.95$\pm$0.04  & 71.77$\pm$8.9& 0.95$\pm$0.03 & 65.89$\pm$8.3 \\ 
     
     & Baseline+TS+ODIN~\cite{liang2018enhancing} & 0.87$\pm$0.05 &45.02$\pm$1.51    & 0.95$\pm$0.05   & 69.42$\pm$2.38 & 0.95$\pm$0.03 & 67.18$\pm$2.16\\ 
     
     & Baseline+energy~\cite{liu2020energy}  & 0.88$\pm$0.03 & 47.77$\pm$1.10 & 0.94$\pm$0.05  & 72.68$\pm$2.69 & 0.96$\pm$0.04   & 74.75$\pm$2.89 \\
     
    & Mixup~\cite{zhang2017mixup} & 0.86$\pm$0.06 &  \bf 53.83$\pm$6.8  &  0.82$\pm$0.11  & 57.01$\pm$8.6 & 0.88$\pm$0.13 &70.51$\pm$9.8 \\ 
     
    AFHQ & DUQ~\cite{van2020uncertainty}       & 0.78$\pm$0.05    & 20.98$\pm$2.0 & 0.67$\pm$0.59  & 16.23$\pm$1.5 & 0.66$\pm$0.55 & 15.34$\pm$2.6  \\ 
     
     & DDU~\cite{mukhoti2021deterministic} & 0.83$\pm$0.02   & 23.19$\pm$2.6 & 0.90$\pm$0.02    &  32.98$\pm$10& 0.75$\pm$0.02 & 10.32$\pm$5.6 \\

    & \bf Baseline+ACE & \bf 0.89$\pm$0.03 &  51.39$\pm$4.4 & \bf 0.98$\pm$0.02    & \bf 88.71$\pm$5.7 &  \bf0.97$\pm$0.03 & \bf 88.87$\pm$9.8 \Bstrut{}\\
    
     \cline{2-8}
    & MC-Dropout~\cite{pmlr-v48-gal16} & 0.84$\pm$0.09  & 30.78$\pm$2.9 & 0.94$\pm$0.02  & 73.59$\pm$2.1& 0.95$\pm$0.02 & 71.23$\pm$1.9 \Tstrut{}\\

    & 5-Ensemble~\cite{NIPS2017_9ef2ed4b} & 0.99$\pm$0.01   & 65.73$\pm$1.2 &   0.97$\pm$0.02  & 89.91$\pm$0.9& 0.99$\pm$0.01 & 92.12$\pm$0.7 \Bstrut{}\\

    \hline
     &  \multicolumn{1}{c}{ }  &   \multicolumn{2}{c}{\bf \nood{Near-OOD} (Digits 7-9) } &  \multicolumn{2}{c}{\bf \food{Far-OOD} (SVHN)} &  \multicolumn{2}{c}{\bf \food{Far-OOD} (fMNIST)} \Tstrut{}\\
     & \bf   & AUC-ROC & TNR@TPR95 & AUC-ROC & TNR@TPR95 &   AUC-ROC & TNR@TPR95 \Bstrut{}\\
    \cline{3-8}
     & Baseline  & 0.86$\pm$0.04  &  28.23$\pm$2.9 &  0.75$\pm$0.15 &  51.98$\pm$0.9 & 0.87$\pm$0.02 & 58.12$\pm$1.5 \Tstrut{}\\  
    
     & Baseline+TS~\cite{guo2017temperaturescaling} & 0.86$\pm$0.01  &  30.12$\pm$2.1 &  0.73$\pm$0.07 &  48.12$\pm$1.5 & 0.89$\pm$0.01 & 61.71$\pm$2.8\\
     
     & Baseline+TS+ODIN~\cite{liang2018enhancing} & 0.83$\pm$0.04 & 34.13$\pm$12.07   & 0.77$\pm$0.13   & 21.59$\pm$19.62 & 0.89$\pm$0.02 & 46.43$\pm$4.31\\ 
     
     & Baseline+energy~\cite{liu2020energy}  & 0.87$\pm$0.04 & \bf 40.30$\pm$1.05  &  0.86$\pm$0.12 & 43.92$\pm$2.30 &  0.91$\pm$0.02  & 62.10$\pm$5.17 \\
     
    Dirty & Mixup~\cite{zhang2017mixup}  & 0.86$\pm$0.02 &  35.46$\pm$1.0 & 0.95$\pm$0.03 &  65.12$\pm$3.1 & 0.94$\pm$0.05 & 66.00$\pm$0.8\\
    
    MNIST & DUQ~\cite{van2020uncertainty}       &  0.78$\pm$0.01   &  15.26$\pm$3.9 & 0.73$\pm$0.03  & 45.23$\pm$1.9 & 0.75$\pm$0.03 & 50.29$\pm$3.1  \\
    
     & DDU~\cite{mukhoti2021deterministic} & 0.67$\pm$0.07   &  10.23$\pm$0.9 &   0.68$\pm$0.04  &  39.31$\pm$2.2 & 0.85$\pm$0.02 & 53.76$\pm$3.7 \\ 
     
     & \bf Baseline+ACE & \bf0.94$\pm$0.02   &   37.23$\pm$1.9   & \bf 0.98$\pm$0.02  &  \bf 67.88$\pm$3.1 & \bf 0.97$\pm$0.02 & \bf 70.71$\pm$1.1 \Bstrut{}\\

    \cline{2-8}
    & MC-Dropout~\cite{pmlr-v48-gal16} & 0.97$\pm$0.02  &  40.89$\pm$1.5 & 0.95$\pm$0.01 &  62.12$\pm$5.7 & 0.93$\pm$0.02 &65.01$\pm$0.7 \Tstrut{}\\  
    
    & 5-Ensemble~\cite{NIPS2017_9ef2ed4b} & 0.98$\pm$0.02  &  42.17$\pm$1.0 &  0.82$\pm$0.03&  55.12$\pm$2.1 & 0.94$\pm$0.01 &64.19$\pm$4.2 \Bstrut{}\\ 
    
    \hline 
    &  \multicolumn{1}{c}{ }  &   \multicolumn{2}{c}{\bf \nood{Near-OOD} (Kids) } &  \multicolumn{2}{c}{\bf \food{Far-OOD} (AFHQ)} &  \multicolumn{2}{c}{\bf \food{Far-OOD} (CIFAR10)} \Tstrut{}\\
     & \bf   & AUC-ROC & TNR@TPR95 & AUC-ROC & TNR@TPR95 &   AUC-ROC & TNR@TPR95 \Bstrut{}\\
     \cline{3-8}
    & Baseline  & 0.84$\pm$0.02 & 1.25$\pm$0.1 & 0.86$\pm$0.03 & 88.57$\pm$0.9 & 0.79$\pm$0.02 & 29.01$\pm$5.1 \Tstrut{}\\ 
    & Baseline+TS~\cite{guo2017temperaturescaling} & 0.82$\pm$0.04 & 1.24$\pm$0.1 & 0.87$\pm$0.06 & 88.75$\pm$0.9 & 0.78$\pm$0.04 & 29.01$\pm$5.1  \\ 
    & Baseline+TS+ODIN~\cite{liang2018enhancing} & 0.65$\pm$0.01 & 8.75$\pm$2.21   &    0.55$\pm$0.01& 23.03$\pm$0.16 & 0.54$\pm$0.01 & 5.00$\pm$0.07\\ 
     & Baseline+energy~\cite{liu2020energy}  & 0.76$\pm$0.51 & 9.40$\pm$0.01 & 0.94$\pm$0.08  & 32.08$\pm$1.70 &  0.85$\pm$0.76  & 17.10$\pm$0.72 \\
    
    & Mixup~\cite{zhang2017mixup} & 0.82$\pm$0.08 & 22.18$\pm$2.7 & 0.95$\pm$0.02 & 82.96$\pm$2.5 & 0.79$\pm$0.13 & 30.54$\pm$1.3\\ 
    CelebA & DUQ~\cite{van2020uncertainty} & 0.80$\pm$0.03 & 14.68$\pm$3.1 & 0.72$\pm$0.07 & 26.62$\pm$7.7 & 0.86$\pm$0.04 & 28.70$\pm$4.1\\ 
    & DDU~\cite{mukhoti2021deterministic} & 0.73$\pm$0.15  & 7.9$\pm$0.4 & 0.74$\pm$0.13  & 8.18$\pm$0.4 & 0.81$\pm$0.15 & 25.45$\pm$1.4 \\ 
    & \bf Baseline+ACE & \bf 0.87$\pm$0.03 & \bf 34.37$\pm$2.5 &    \bf 0.96$\pm$0.01 & \bf 96.35$\pm$2.5 & \bf 0.92$\pm$0.05 &  \bf 63.51$\pm$1.5  \Bstrut{}\\
    \cline{2-8}
    & MC-Dropout~\cite{pmlr-v48-gal16} & 0.70$\pm$0.10 & 25.62$\pm$1.7 & 0.86$\pm$0.1 & 91.72$\pm$7.5 & 0.74$\pm$0.12 & 64.79$\pm$1.8 \Tstrut{}\\ 
    & 5-Ensemble~\cite{NIPS2017_9ef2ed4b} & 0.93$\pm$0.03 & 10.35$\pm$0.2 & 0.99$\pm$0.0 & 98.31$\pm$1.2 & 0.92$\pm$0.10 & 61.88$\pm$1.2 \Bstrut{}\\ 
    
    \hline
      &  \multicolumn{1}{c}{ }  &   \multicolumn{2}{c}{\bf \nood{Near-OOD} (other lesions) } &  \multicolumn{2}{c}{\bf \food{Far-OOD} (CelebA)} &  \multicolumn{2}{c}{\bf \food{Far-OOD} (Skin-texture)} \Tstrut{}\\
     & \bf   & AUC-ROC & TNR@TPR95 & AUC-ROC & TNR@TPR95 &   AUC-ROC & TNR@TPR95 \Bstrut{}\\
    \cline{3-8}
     & Baseline  &  0.67$\pm$0.04  &8.70$\pm$2.5    &  0.66$\pm$0.06  & 10.00$\pm$3.6 &  0.65$\pm$0.10 & 5.91$\pm$2.8 \Tstrut{}\\ 
     
      & Baseline+TS~\cite{guo2017temperaturescaling} & 0.67$\pm$0.05   & 8.69$\pm$2.0 & 0.63$\pm$0.06  & 9.24$\pm$4.3 & 0.68$\pm$0.07 & 5.70$\pm$3.2 \\ 
      & Baseline+TS+ODIN~\cite{liang2018enhancing} & 0.68$\pm$0.01 & 9.43$\pm$0.33   & 0.67$\pm$0.07   & 11.32$\pm$4.66 & 0.68$\pm$0.07 & 6.60$\pm$0.29\\ 
      & Baseline+energy~\cite{liu2020energy}  & 0.70$\pm$0.04 & 10.85$\pm$0.08 & 0.70$\pm$0.14  & 7.90$\pm$0.29 & 0.65$\pm$0.20   & 2.83$\pm$1.33 \\

    Skin& Mixup~\cite{zhang2017mixup} & 0.67$\pm$0.01 & 8.52$\pm$2.8   & 0.64$\pm$0.08  & 10.21$\pm$4.0 &  0.72$\pm$0.05 &5.26$\pm$3.1 \\ 
    
    Lesion & DUQ~\cite{van2020uncertainty} & 0.67$\pm$0.04  &   3.12$\pm$1.8 & 0.89$\pm$0.09 &   11.89$\pm$2.5 & 0.64$\pm$0.03 & 4.8$\pm$1.5\\
    
     & DDU~\cite{mukhoti2021deterministic} & 0.65$\pm$0.03  &   3.45$\pm$1.9 & 0.75$\pm$0.04 &  15.45$\pm$2.9 & 0.71$\pm$0.05 & 4.19$\pm$1.3 \\
    
    & \bf Baseline+ACE & \bf0.72$\pm$0.04   & \bf 10.99$\pm$2.8 & \bf 0.97$\pm$0.02   & \bf 66.77$\pm$1.4 &  \bf 0.96$\pm$0.03 & \bf 95.83$\pm$5.0 \Bstrut{} \\
     
     \cline{2-8}
    & MC-Dropout~\cite{pmlr-v48-gal16} & 0.67$\pm$0.05   & 9.45$\pm$3.9 &  0.80$\pm$0.07  & 30.00$\pm$3.2&0.56$\pm$0.03  & 10.87$\pm$2.3 \Tstrut{}\\

    & 5-Ensemble~\cite{NIPS2017_9ef2ed4b} & 0.88$\pm$0.01   & 11.23$\pm$1.7 &  0.91$\pm$0.03  & 27.89$\pm$5.9 & 0.76$\pm$0.02 & 17.89$\pm$3.5 \Bstrut{}\\

    \hline
    \end{tabular}}
\end{center}
\end{table*}

 In Fig.~\ref{fig:all}, we observe much overlap between the PE distribution of the near-OOD samples  and in-distribution samples in Fig.~\ref{fig:all}.C. Further, in Fig.~\ref{fig:all}.D, we see that our model successfully disentangles OOD samples from the in-distribution samples by using density estimates from the discriminator of the PCE. In Table~\ref{OOD-table}, we report the AUC-ROC and TNR@TPR95 scores on detecting the two types of OOD samples. We first use the discriminator from the PCE to detect far-OOD samples. The discriminator achieved near-perfect AUC-ROC for detecting far-OOD samples. We used the PE estimates from the fine-tuned model (baseline + ACE) to detect near-OOD samples. Overall our model outperformed other methods on both near and far-OOD detection tasks with high TNR@TPR95.

\subsection{Robustness to Adversarial Attacks}

We compared the baseline model before and after fine-tuning (baseline + ACE) in their robustness to three adversarial attacks: Fast Gradient Sign Method (FGSM)~\cite{fgsm_attack}, Carlini-Wagner (CW)~\cite{carlini_wagner_attack}, and DeepFool~\cite{deepfool_attack}. 

 \begin{figure}[ht]
      \includegraphics[width=1.0\linewidth]{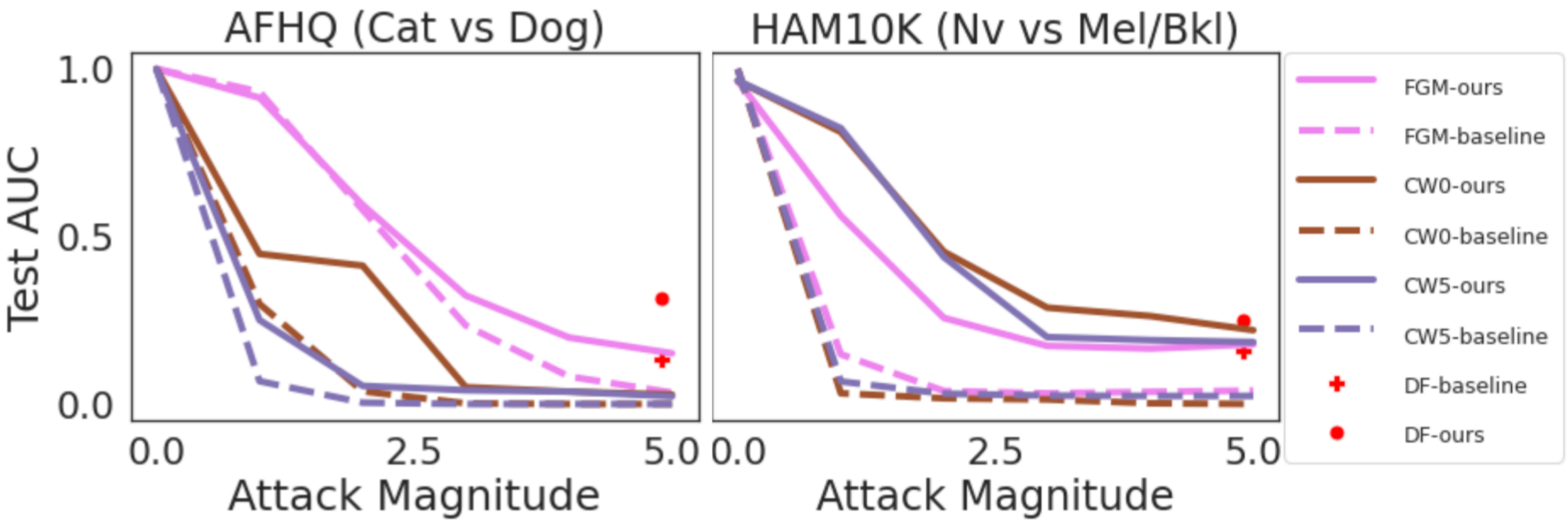}
    \caption{\footnotesize Plots comparing baseline model before and after fine-tuning (ACE) for different magnitudes of adversarial attack. The figure shows three different attacks -- FGSM~\cite{fgsm_attack}, CW~\cite{carlini_wagner_attack}, DeepFool~\cite{deepfool_attack}, on three different datasets -- HAM10K, AFHQ, MNIST. The x-axis denotes maximum perturbation ($\epsilon$) for FGSM, and iterations in multiples of $10$ for CW and DeepFool. Attack magnitude of $0$ indicates no attack. For CW we used $\kappa=0$ and $5$. (All results are reported on the test-set of the classifier).}\label{fig:adv_attacks}
  \end{figure}

For each attack setting, we transformed the test set into an adversarial set.  In Fig.~\ref{fig:adv_attacks}, we report the AUC-ROC over the adversarial set as we gradually increase the magnitude of the attack. For FGSM,  we use the maximum perturbation ($\epsilon$) to specify the attack's magnitude. For CW, we gradually increase the number of iterations to an achieve a higher magnitude attack. We set box-constraint parameter as $c=1$, learning rate $\alpha=0.01$ and confidence $\kappa=0,5$. For  DeepFool ($\eta=0.02$), we show results on the best attack. Our improved model (baseline + ACE) consistently out-performed the baseline model in test AUC-ROC, thus showing an improved robustness to all three attacks.


\section{Conclusion}

We propose a novel application of counterfactual explanations in improving  the uncertainty quantification of a  \emph{pre-trained} DNN. We improved upon the existing work on counterfactual explanations, by proposing a StyleGANv2-based backbone. Fine-tuning on augmented data, with soft labels helps in improving the decision boundary and the fine-tuned model, combined with the discriminator of the PCE can successfully capture uncertainty over ambiguous samples, unseen near-OOD samples with label shift and far-OOD samples from independent datasets. We out-performed state-of-the-art methods for uncertainty quantification on four datasets, and our improved model also exhibits robustness to adversarial attacks.


{\small
\bibliographystyle{ieee_fullname}
\bibliography{egbib}
}

\section{Supplemental Material}
\subsection{Implementation Details}
\subsubsection{Dataset}
We focus on improving classification models based on deep convolution neural networks (CNN) as most state-of-the-art performance models fall in this regime. In our experiments, we consider classification models trained on following datasets:

 \begin{figure*}[htbp]
  \centering
  \includegraphics[width=0.9\linewidth]{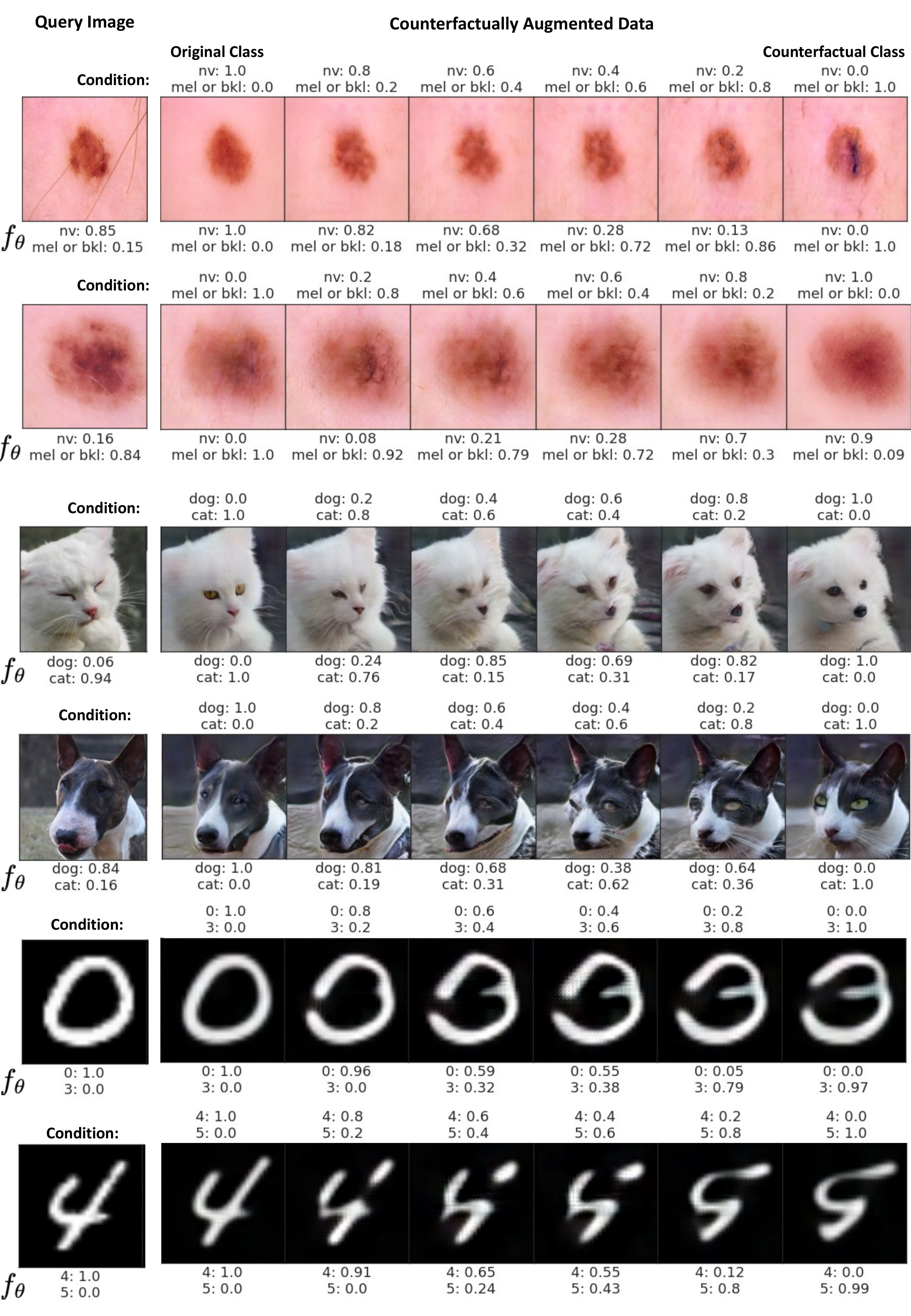}
  \footnotesize
  \caption {Examples of data augmentation using counterfactual explanations for different datasets.}
  \label{fig:example}
\end{figure*} 

\begin{figure*}[h]
  \centering
  \includegraphics[width=0.8\linewidth]{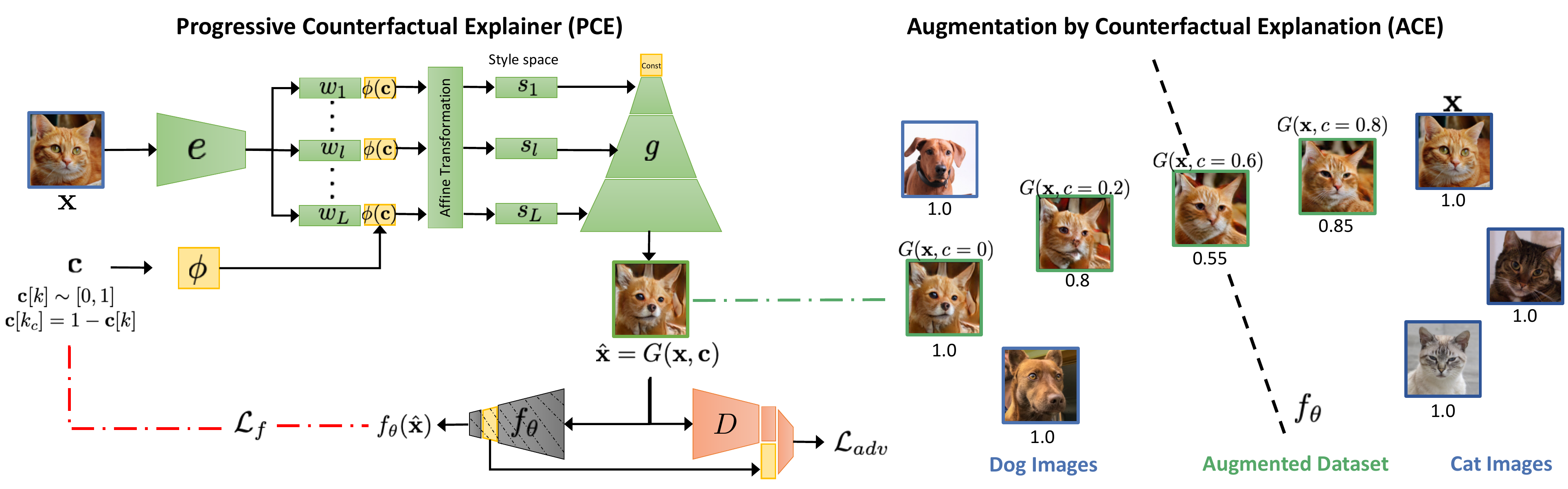}
  \footnotesize
  \caption{PCE: The encoder-decoder architecture to create counterfactual augmentation for a given query image. ACE: Given a query image, the trained PCE generates a series of perturbations that gradually traverse the decision boundary of $f_{\theta}$ from the original class to a counter-factual class,  while still remaining plausible and realistic-looking.    }
  \label{fig:PCE_SM}
\end{figure*} 

\begin{enumerate}
\item AFHQ~\cite{choi2020afhq}: Animal face high quality (AFHQ) dataset is a high resolution dataset of animal faces with 16K images from cat, dog and wild labels. In our experiments, we consider a multi-class classifier over cat and dog labels. We consider images with ``wild" label as near-OOD. The classifier is trained at an image resolution of $256\times256$.

\item Dirty MNIST~\cite{mukhoti2021deterministic}: The dataset is a combination of original MNIST~\cite{lecun1998mnist} and simulated Ambiguous-MNIST dataset. Each sample in Ambiguous-MNIST is constructed by decoding a linear combination of latent representations of two different MNIST digits from a pre-trained VAE~\cite{kingma2014autoencoding}. The training dataset of the classifier comprises of 60K clean-MNIST and 60K Ambiguous-MNIST samples, with one-hot labels. In our experiments, we consider classifier trained on seven classes over digits `0' - `6'. We consider images from digits `7' - `9' as near-OOD samples. The original dataset consists of grayscale images of size 28$\times$28 pixels. We consider a classification model trained on 64$\times$64 resolution.
     
 \item Skin lesion (HAM10K)~\cite{tschandl2018ham10000}:  The HAM10000 is a dataset of 100K dermatoscopic images of pigmented skin lesions. It contains seven different lesion types -- Melanocytic Nevi (nv), Melanoma (mel), Benign Keratosis (bkl), Actinic Keratoses and Intraepithelial Carcinoma (akiec), Basal Cell Carcinoma (bcc), Dermatofibroma (df), Vascular skin lesions (vasc). In our experiments, we consider classifier trained to distinguish the majority class nv from mel and bkl. We consider images from rest of the lesions as near-OOD.  The classifier is trained at an image resolution of $256\times256$.

 \item CelebA~\cite{liu2015celeba}
 : Celeb Faces Attributes Dataset (CelebA) is a large-scale face attributes dataset with more than 200K celebrity images, each with 40 binary attributes annotations per image. In our experiments, we consider a two-class classifier over attributes ``Young" and ``Smiling" trained on CelebA dataset. Our \aid{AiD} samples comprises of middle-aged people who are arguably neither young nor old. To obtain such data, we use aleatoric uncertainty estimates from MC-Dropout averaged across 50 runs on test-set of CelebA. The classifier is trained at an image resolution of $256\times256$. We center-crop the images as a pre-processing step.

 \end{enumerate}

  \subsubsection{Classification Model}
We used DenseNet architecture as the classification model.  In DenseNet, each layer implements a non-linear transformation based on composite functions such as Batch Normalization (BN), rectified linear unit (ReLU), pooling, or convolution. The resulting feature map at each layer is used as input for all the subsequent layers, leading to a highly convoluted multi-level multi-layer non-linear convolutional neural network.  We aim to improve such a model in a post-hoc manner without accessing the parameters learned by any layer or knowing the architectural details. Our proposed approach can be used for any DNN architecture.

 \subsection{Progressive Counterfactual Explainer}

We formulate the progressive counterfactual explainer (PCE) as a composite of two functions, an image encoder $e(\cdot)$ and a conditional decoder ($g(\cdot)$)~\cite{abdal2019image2stylegan}. Our architecture for the conditional decoder is adapted from StyleGANv2~\cite{abdal2019image2stylegan}. The image encoder converts the input image $\mathbf{x}$ into $l$ different latent codes ($w_l \in \mathbb{R}^{512}$), for each of the $L$ layers of the decoder. The decoder further transforms the layer-specific latent representation into a layer-specific style-vector as $s_l = A_l([w_l , \phi(\mathbf{c})])$ where, $A_l$ is an affine transformation and $\phi(\mathbf{c})$ is an embedding for $\mathbf{c}$. For training the StyleGANv2 decoder, we consider the default training parameters from~\cite{abdal2019image2stylegan}. For training the  PCE, we use a randomly sampled subset $(\sim50\%)$ of the baseline training data. Given an input image, the predicted class $k$ and a counterfactual class $k_c$, we initialize the condition $\mathbf{c}$ with all zeros and then set $\mathbf{c}[k] \sim \text{Uniform}(0,1) $ and $\mathbf{c}[k_c] = 1 -  \mathbf{c}[k]$. In all our experiments, we used $\lambda_{adv} = 10$, $\lambda_{rec} = 100$ and $\lambda_{f} = 10$. Fig.~\ref{fig:PCE_SM} summarizes our architecture.

For generating counterfactually augmented data, we first consider a randomly selected subset of real training data as $\mathcal{X}_r \in \mathcal{X}$. For each image in $\mathcal{X}_r$, we generate four augmented images by randomly selecting the $\mathbf{c}[k]$. For each augmented image, we used the condition used to generate the image as the soft label while fine-tuning. Fig.~\ref{fig:PCE_SM} shows an example of our data augmentation. We denote the pool of the augmented images as $\mathcal{X}_c$. In Fig.~\ref{fig:example}, we show examples of counterfactual augmentation from different datasets.

\begin{figure*}[h]
  \centering
  \includegraphics[width=0.8\linewidth]{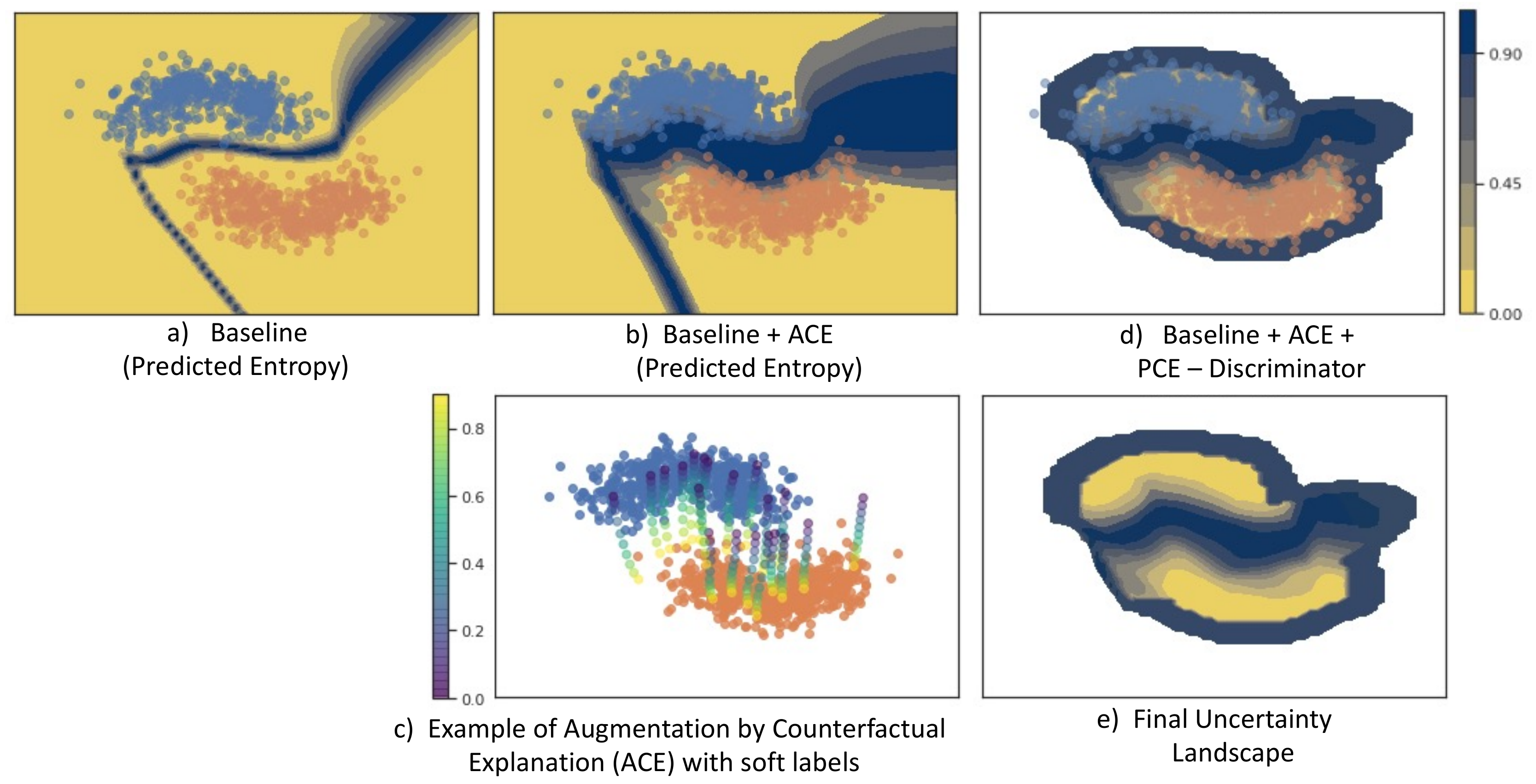}
  \caption { \footnotesize Uncertainty results on Two Moons dataset. Yellow indicates low uncertainty, while blue indicates uncertainty. a) The baseline classifier is uncertain only along the decision boundary, and certain elsewhere. b) Fine-tuning baseline model on ACE data  improves uncertainty estimates near the decision boundary. c) An example of augmented data and corresponding soft labels. d) The discriminator from PCE rejects OOD samples, hence the rejected space have no uncertainty values (white color). e) The final uncertainty landscape, the improved classifier is certain on in in-distribution regions and rejects OOD data.    }
  \label{fig:2D}
\end{figure*}

For fine-tuning the given baseline with consider a combination of the original training dataset $\mathcal{X}$ and the augmented data $\mathcal{X}_c$. We randomly selected a subset of samples from the two distributions and fine-tune the baseline for 5 to 10 epochs. We used the expected calibration error and the test-set accuracy to choose the final checkpoint. Our model does not require access to OOD or AiD dataset during fine-tuning. During evaluation we compute predicted entropy (PE) for original test-set and OOD samples and   measure for a range of thresholds how well
the two are separated. We report the AUC-ROC and the true negative rate (TNR) at 95\% true positive rate (TPR) (TNR@TPR95) in our results (\emph{see Table~1~and~2}). 

\subsection{Toy-Setup: Two-Moons}
In this section, we demonstrate our method on a toy setup: the Two Moons dataset.
We used the experimental set-up from DDU~\cite{mukhoti2021deterministic} for this experiment. We use scikit-learn’s datasets package to generate 2000 samples with a noise rate of 0.1. For baseline classification model, we use a 2-layer dense-layer architecture, with ReLU activation and batch normalization. The 2-D input data is projected to a 64-D latent space and then to 1D space to make final binary prediction. In Fig.~\ref{fig:2D}.a, we show the uncertainty estimates (predicted entropy PE) from the baseline classifier. The baseline classifier is uncertain only along the decision boundary, and certain elsewhere (low PE).

Given the baseline classifier, we train a PCE to generate augmented data. We use an encoder with two fully-connected layers that map 2-D input data to a 64-D latent space. The condition is also projected to a 64-D space and is concatenated to the output of the encoder. The decoder also have two fully-connected layers that maps the concatenated  128-D latent vector back to a 2-D input space. In Fig.~\ref{fig:2D}.c, we show example of augmentation by counterfactual explanation (ACE). Given a query point, we generate series of augmented data by gradually changing the condition such that the decision of the baseline is flipped. The color of the dot represents the conditioned used to create the augmented sample. Next, we fine-tune the classifier using the original and the counterfactually augmented data. In Fig.~\ref{fig:2D}.b, we show the PE estimates from the fine-tuned classifier (baseline + ACE). Fine-tuning with our augmented data widen the decision boundary. Finally, we used the discriminator of the PCE as a density estimator, to identify and reject OOD data. The discriminator is trained on real/fake samples near the training distribution. Hence, we used a threshold of 0.5 on the discriminator to reject everything that is far from the training distribution. In Fig.~\ref{fig:2D}.d, the white color show the input space that is rejected by the discriminator.  In Fig.~\ref{fig:2D}.e, we show the final uncertainty landscape without overlaying the training data. We improved the baseline model, to have high certainty only in in-distribution regions. The uncertainty increases as we go near the decision boundary. Thus in addition to image classifiers, our strategy improves the uncertainty estimates even for a classifier trained on a small 2D setup like Two Moons. 

\subsection{Additional Results}

\begin{table*}[!ht]
\caption{ \footnotesize Additional results on identifying \textbf{\aid{ambiguous in-distribution (AiD)}} samples. For all metrics, higher is better.}
\label{aid_table2}
\begin{center}
\footnotesize
    \begin{tabular}{cccccc}
    \hline
    \multicolumn{1}{c}{\bf Train }  &  \multicolumn{1}{c}{\bf Method/ }  & \multicolumn{1}{c}{\bf Test-Set} &  \multicolumn{2}{c}{\bf Identifying \aid{AiD}}  \Tstrut{}\\
    \bf Dataset & \bf Model  & Accuracy & AUC-ROC & TNR@TPR95  \Bstrut{}\\
    \hline
    & Baseline+energy~\cite{liu2020energy}  &   99.44$\pm$0.02  &  0.87$\pm$0.06  & 49.00$\pm$1.64 \Tstrut{}\\
    AFHQ & Energy w/ fine-tune~\cite{liu2020energy}  &  99.45$\pm$0.11   & 0.69$\pm$1.28   & 30.36$\pm$2.52 \\
    & Outlier Exposure~\cite{hendrycks2018deep}  &  99.50$\pm$0.14   &  0.85$\pm$0.01  & 41.07$\pm$0.75 \\ 
    & \bf Baseline+ACE  &  \bf 99.52$\pm$0.21   & \bf 0.91$\pm$0.02   & \bf 50.75$\pm$3.9 \Bstrut{}\\ 
    \hline 
    & Baseline+energy~\cite{liu2020energy}  &  95.68$\pm$0.02   & 0.80$\pm$0.03   & 17.60$\pm$0.55 \Tstrut{}\\ 
    Dirty & Energy w/ fine-tune~\cite{liu2020energy}  & 96.17$\pm$0.02    &  0.39$\pm$0.04  & 11.59$\pm$0.25 \\
    MNIST & Outlier Exposure~\cite{hendrycks2018deep}  &  \bf 96.30$\pm$0.07   & 0.63$\pm$0.07   & 17.6$\pm$2.88 \\ 
    & \bf Baseline+ACE  &  95.36$\pm$0.45   &\bf 0.86$\pm$0.01   & \bf 34.12$\pm$2.60 \Bstrut{}\\ 
    \hline
    & Baseline+energy~\cite{liu2020energy}  &  89.36$\pm$0.96   & 0.57$\pm$0.28   & 4.87$\pm$0.32 \Tstrut{}\\ 
    CelebA & Energy w/ fine-tune~\cite{liu2020energy}  & \bf 90.22$\pm$0.96   & 0.53$\pm$1.25   & 5.06$\pm$0.28 \\
    & Outlier Exposure~\cite{hendrycks2018deep}  &   86.65$\pm$1.22  &  0.53$\pm$0.46  & 5.06$\pm$0.19 \\ 
    & \bf Baseline+ACE  &   86.80$\pm$0.79  & \bf 0.74$\pm$0.06   & \bf 22.36$\pm$2.30 \Bstrut{}\\ 
    \hline
    & Baseline+energy~\cite{liu2020energy}  & 85.88$\pm$0.75    & 0.77$\pm$0.12   & 18.40$\pm$0.51 \Tstrut{}\\ 
    Skin-Lesion & Energy w/ fine-tune~\cite{liu2020energy}  &  \bf 86.56$\pm$0.53   &0.64$\pm$0.06    & 17.45$\pm$1.78 \\
    (HAM10K)& Outlier Exposure~\cite{hendrycks2018deep}  &  86.37$\pm$0.46   & 0.73$\pm$0.02   & 13.21$\pm$2.70 \\ 
    & \bf Baseline + ACE  &  81.21$\pm$1.12   & \bf 0.84$\pm$0.05   & \bf 71.60$\pm$3.80 \Bstrut{}\\ 
    \hline
    \end{tabular}
\end{center}
\end{table*}

Much of the prior work has focused on obtaining uncertainty estimates from a pre-trained DNN output using threshold-based scoring functions. Liu~et~al~\cite{liu2020energy} in their paper show how energy functions can be used not only as scoring functions but also as a trainable cost-function to shape the energy surface explicitly for OOD-detection. Hendrycks~et~al~\cite{hendrycks2018deep} propose the Outlier Exposure method which regularizes the softmax outputs to be a uniform distribution for outlier data. We compare these commonly-used methods against our technique (ACE) and show the results in Tables~\ref{aid_table2}~and~\ref{ood_table2}. Our method is consistent and competitive, if not outperforming, across all datasets and AiD/OOD categories.

 \begin{table*}[!htb]
\caption{ \footnotesize OOD detection performance for different scoring-based methods.}
\label{ood_table2}
\begin{center}
\scriptsize
\resizebox{\textwidth}{!}{%
    \begin{tabular}{cccccccc}
    \hline 
    \multicolumn{1}{c}{\bf Train }  &  \multicolumn{1}{c}{\bf Method }  &   \multicolumn{2}{c}{\bf \nood{Near-OOD} (Wild)} &  \multicolumn{2}{c}{\bf \food{Far-OOD} (CIFAR10)} &  \multicolumn{2}{c}{\bf \food{Far-OOD} (CelebA)} \Tstrut{}\\
    \bf Dataset & \bf   & AUC-ROC & TNR@TPR95 & AUC-ROC & TNR@TPR95 &   AUC-ROC & TNR@TPR95 \Bstrut{}\\
    \hline
     & Baseline+energy~\cite{liu2020energy}  & 0.88$\pm$0.03 & 47.77$\pm$1.10 & 0.94$\pm$0.05  & 72.68$\pm$2.69 & 0.96$\pm$0.04   & 74.75$\pm$2.89 \Tstrut{}\\
    AFHQ & Energy w/ fine-tune~\cite{liu2020energy}  &    \bf 0.93$\pm$3.06 & 45.97$\pm$2.78 & \bf 0.99$\pm$0.00  & 0.66$\pm$0.01 & 0.94$\pm$1.86 & 68.38$\pm$3.03  \\ 
     & Outlier Exposure~\cite{hendrycks2018deep} & 0.92$\pm$0.01  & \bf 73.99$\pm$2.62 & 0.99$\pm$0.20 & \bf 99.54$\pm$0.79 & 0.96$\pm$0.01 & 78.69$\pm$3.02\\ 
    & \bf Baseline+ACE & 0.89$\pm$0.03 & 51.39$\pm$4.40 & 0.98$\pm$0.02  & 88.71$\pm$5.70 & \bf 0.97$\pm$0.03  & \bf 88.87$\pm$9.80 \Bstrut{}\\
    \hline
     & Baseline+energy~\cite{liu2020energy}  & 0.87$\pm$0.04 & 40.30$\pm$1.05  &  0.86$\pm$0.12 & 43.92$\pm$2.30 &  0.91$\pm$0.02  & 62.10$\pm$5.17 \Tstrut{}\\ 
    Dirty & Energy w/ fine-tune~\cite{liu2020energy}  & 0.60$\pm$0.08 & 37.43$\pm$0.93 & \bf 1.00$\pm$0.00  & \bf 99.99$\pm$0.00 & \bf 1.00$\pm$0.00 & 99.06$\pm$0.01  \\ 
    MNIST & Outlier Exposure~\cite{hendrycks2018deep} &  \bf 0.94$\pm$0.01 & \bf 65.58$\pm$1.64 & \bf 1.00$\pm$0.00 & \bf 99.99$\pm$0.00 & \bf 1.00$\pm$0.00 & \bf 99.56$\pm$0.12 \\ 
    & \bf Baseline+ACE & \bf 0.94$\pm$0.02 & 37.23$\pm$1.90 & 0.98$\pm$0.02  & 67.88$\pm$3.10 & 0.97$\pm$0.02  & 70.71$\pm$1.10 \Bstrut{}\\
    \hline
     & Baseline+energy~\cite{liu2020energy}  & 0.76$\pm$0.51 & 9.40$\pm$0.01 & 0.94$\pm$0.08  & 32.08$\pm$1.70 &  0.85$\pm$0.76  & 17.10$\pm$0.72 \Tstrut{}\\
    CelebA& Energy w/ fine-tune~\cite{liu2020energy}  & 0.85$\pm$1.27    & 32.81$\pm$1.92 & \bf 0.99$\pm$0.00  & \bf 99.99$\pm$0.00 & 0.91$\pm$0.77 & \bf 84.35$\pm$1.29  \\ 
     & Outlier Exposure~\cite{hendrycks2018deep} & 0.66$\pm$0.69  & 8.44$\pm$0.45 & 0.75$\pm$0.70 & 26.09$\pm$0.51 & 0.69$\pm$0.53 & 16.63$\pm$0.90\\ 
    & \bf Baseline+ACE & \bf 0.87$\pm$0.03 & \bf 34.37$\pm$2.50 & 0.96$\pm$0.01  & 96.35$\pm$2.50 & \bf 0.92$\pm$0.05  & 63.51$\pm$1.50  \Bstrut{}\\
    \hline
     & Baseline+energy~\cite{liu2020energy}  & 0.70$\pm$0.04 & 10.85$\pm$0.08 & 0.70$\pm$0.14  & 7.90$\pm$0.29 & 0.65$\pm$0.20   & 2.83$\pm$1.33 \Tstrut{}\\
    Skin-Lesion & Energy w/ fine-tune~\cite{liu2020energy}  & 0.62$\pm$0.02 & 9.80$\pm$1.81 &  \bf 1.00$\pm$0.00 & \bf 99.77$\pm$0.33 & 0.76$\pm$0.13 & 16.04$\pm$1.08  \\ 
    (HAM10K) & Outlier Exposure~\cite{hendrycks2018deep} & 0.67$\pm$0.09 & 10.38$\pm$3.30 & 0.99$\pm$0.00 & 97.17$\pm$2.37 & 0.81$\pm$0.08 & 22.64$\pm$4.30 \\ 
    & \bf Baseline+ACE & \bf 0.72$\pm$0.04 & \bf 10.99$\pm$2.80 & 0.97$\pm$0.02  & 66.77$\pm$1.40 & \bf 0.96$\pm$0.03  & \bf 95.83$\pm$5.00 \Bstrut{}\\
    \hline 
    \end{tabular}}
\end{center}
\end{table*}

\subsection{Ablation Study}
 \begin{figure*}[!htb]
  \centering
  \includegraphics[width=0.83\linewidth]{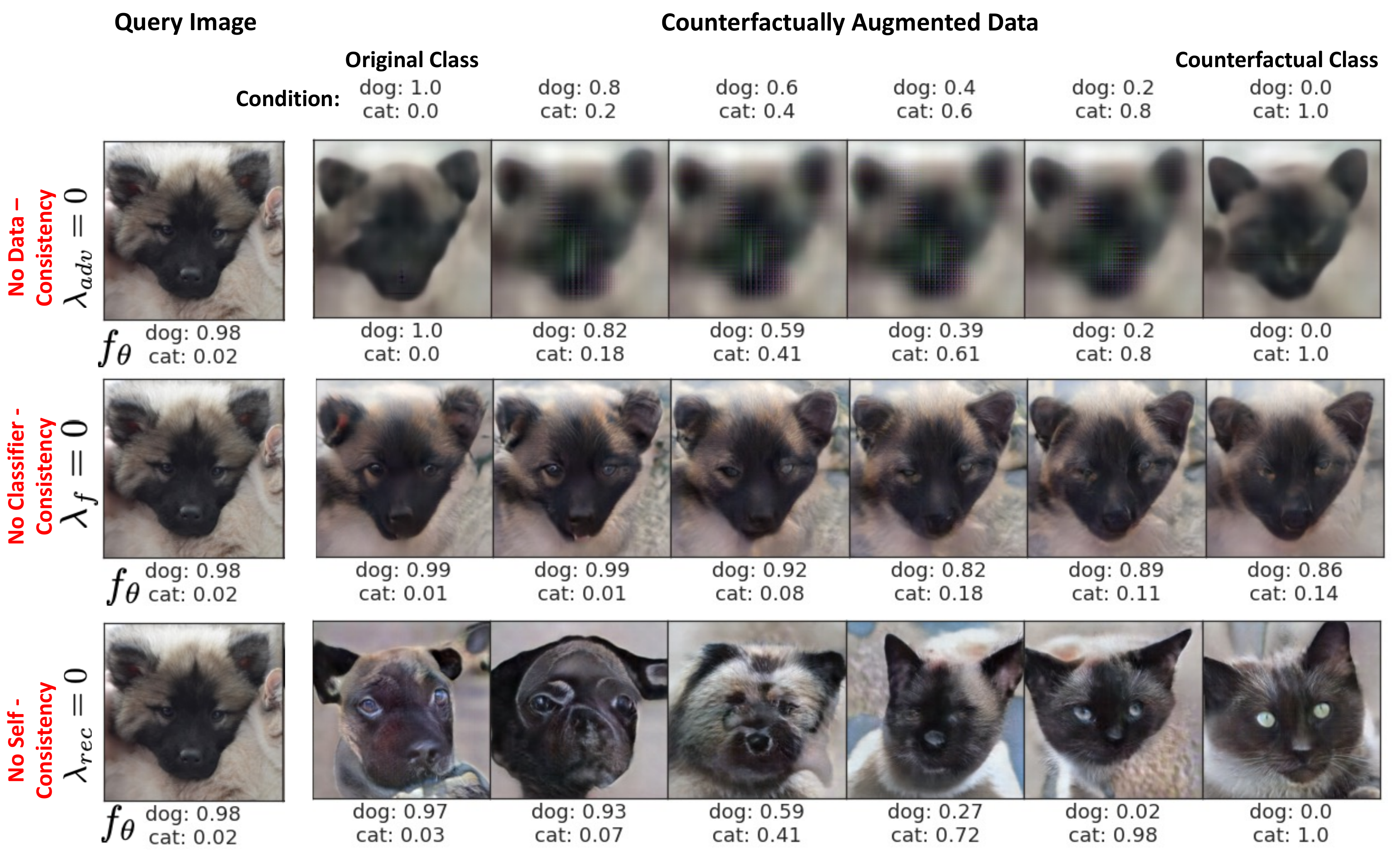}
  \footnotesize
  \caption {Examples of data augmentation while ablating different loss terms.}
  \label{fig:ablation}
\end{figure*} 

 \begin{figure*}[!htb]
  \centering
  \includegraphics[width=0.85\linewidth]{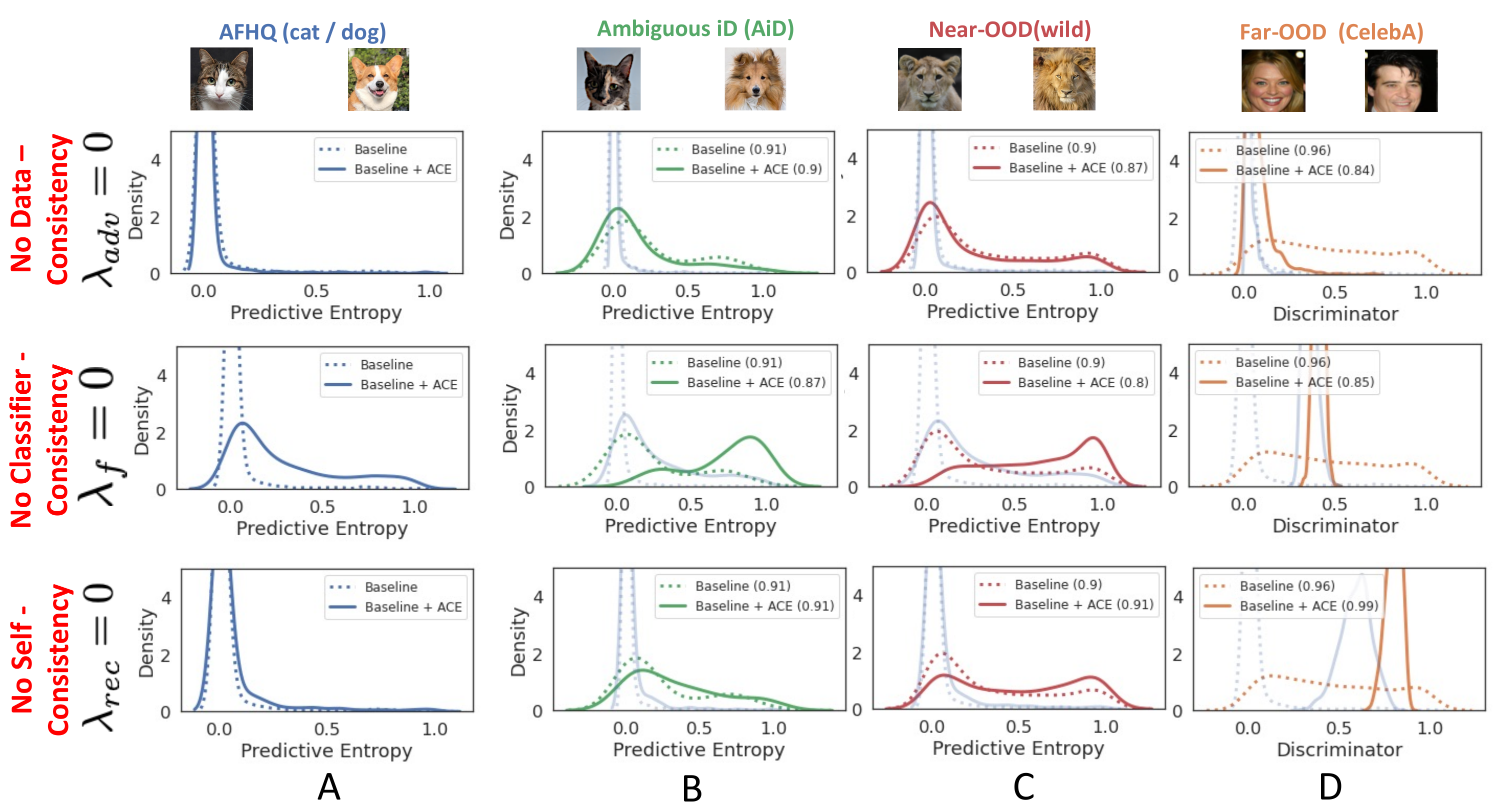}
  \footnotesize
  \caption {Comparison of the uncertainty estimates from the baseline, before and after the fine-tuning with ACE. Each row represents a different ablation over the three loss terms. A) Predicted entropy (PE) of \textbf{\uid{in-distribution (iD)}} samples. Ideally, fine-tuning should minimally effect the PE distribution over iD samples. Without classification consistency loss (second row), the PE distribution of iD samples changed significantly. B)  PE distribution over \textbf{\aid{ ambiguous in-distribution (AiD)}} samples. C) PE distribution over \textbf{\nood{near-OOD}} samples.   The data augmentation derived from PCE without adversarial loss or reconstruction loss, is not able to separate AiD samples or near-OOD from rest of the test set. D)  We use the discriminator of the PCE to identify \textbf{\food{far-OOD}} samples. In all three rows, we observe sub-optimal performance of the discriminator in identifying and rejecting far-OOD samples.  The legend shows the AUC-ROC for binary classification over uncertain samples and iD samples. Hence, all three loss terms are important to improve the uncertainty estimates of the baseline over all samples across the uncertainty spectrum.  }
  \label{fig:ablation_all}
\end{figure*}

We conducted an ablation study over the three loss terms of PCE in Eq.~5. The three terms of the loss function enforces three properties of counterfactual explanation, data consistency: explanations should be realistic looking images, classifier consistency: explanations should produce a desired outcome from the classifier and self consistency: explanation image should retain the identity of the query image. For ablation study we consider the cat and dog classifier. We train three PCE, in each run we ablate one term from the final loss function. In Fig.~\ref{fig:ablation}, we show qualitative example of the counterfactual  data augmentation generated through each PCE. Without data consistency, the images are blur and are no longer realistic. Without classifier consistency loss, though the images are realistic, but the output of the classifier is not changing with the condition, hence such PCE won't generate augmented samples near the decision boundary, which is the goal of our proposed strategy. With self consistency, the generated images are not a gradual transformation of a given query image. 

Further, in Fig.~\ref{fig:ablation_all} we present quantitatively compare  the uncertainty estimates from the baseline, before and after the fine-tuning with ACE. In each row, we represent a different ablation over the three loss terms. Fig.~\ref{fig:ablation_all}.A. shows the predicted entropy (PE) of \textbf{\uid{in-distribution (iD)}} samples. Ideally, fine-tuning should minimally effect the PE distribution over iD samples. Without classification consistency loss (second row), the PE distribution of iD samples changed significantly. Fig.~\ref{fig:ablation_all}.B and  Fig.~\ref{fig:ablation_all}.C shows the PE distribution over \textbf{\aid{ ambiguous in-distribution (AiD)}} samples and  \textbf{\nood{near-OOD}} samples, respectively.   The data augmentation derived from PCE without adversarial loss or reconstruction loss, is not able to separate AiD samples or near-OOD from rest of the test set. In Fig.~\ref{fig:ablation_all}.D, we use the discriminator of the PCE to identify \textbf{\food{far-OOD}} samples. In all three rows, we observe sub-optimal performance of the discriminator in identifying and rejecting far-OOD samples.  The legend shows the AUC-ROC for binary classification over uncertain samples and iD samples. Hence, all three loss terms are important to improve the uncertainty estimates of the baseline over all samples across the uncertainty spectrum.


\subsection{Robust Generalization}
In this experiment, we establish a connection between loss landscape plots and generalization of classifiers. In order to qualitatively understand the improved generalization of our method, we try to visualize the high-dimensional loss landscape via 3D weight visualization plots as shown by Li \etal. We compute the cross-entropy loss using test set of CelebA and AFHQ and follow the method given by Li \etal to compare the loss landscape geometry for the baseline model and our method (ACE). 

We observe that our method leads to smooth and flatter loss landscapes as compared to baseline. This shows that slight perturbation to the weight does not change the loss much, which may qualitatively explain why we obtain better generalization performance and robustness to adversarial attacks in our experiments. We do not thoroughly investigate this direction and leave it as an important direction for future research.

 \begin{figure}[!htbp]
\centering
  \includegraphics[width=0.7\linewidth]{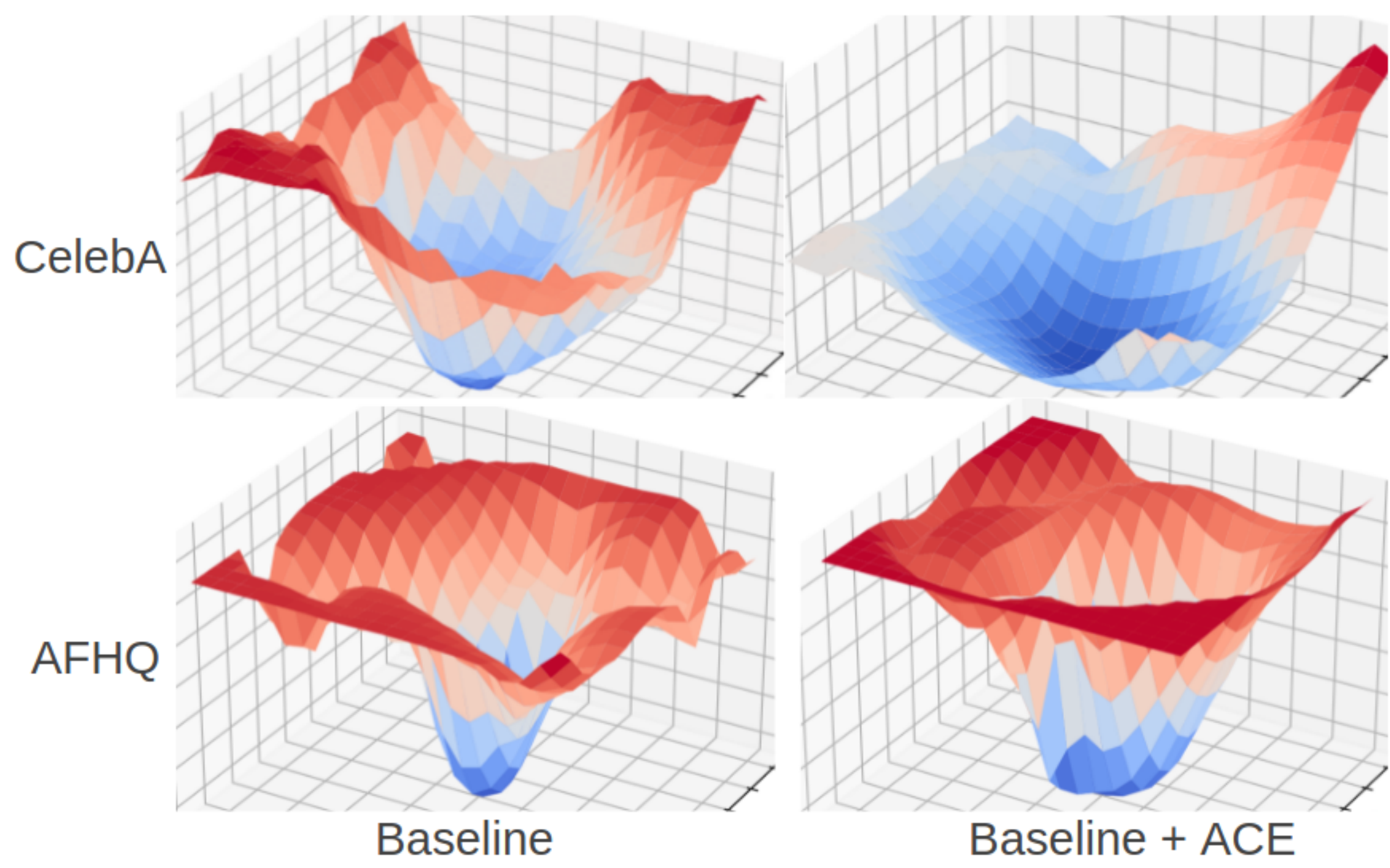}
\caption{\footnotesize Weight loss landscape visualizations for baseline model and our method on CelebA and AFHQ datsets}
\label{fig:celeba_afhq_landscape}
\end{figure}

\end{document}